\documentclass[transmag]{IEEEtran}

\usepackage{algorithm}
\usepackage{multirow}
\usepackage{makecell}
\usepackage{amsfonts}
\usepackage{booktabs}
\usepackage{color}

\ifCLASSINFOpdf
  \usepackage[pdftex]{graphicx}
\else
  \usepackage[dvips]{graphicx}
  \DeclareGraphicsExtensions{.eps,.jpeg}
\fi
\usepackage[cmex10]{amsmath}
\usepackage{algorithmic}
\ifCLASSOPTIONcompsoc
 \usepackage[caption=false,font=normalsize,labelfont=sf,textfont=sf]{subfig}
\else
 \usepackage[caption=false,font=footnotesize]{subfig}
\fi
\usepackage{url}

\begin{document}
%
\title{Scalable Image Retrieval by Sparse Product Quantization}

\author{Qingqun~Ning,
Jianke~Zhu,~\IEEEmembership{Member,~IEEE,}
Zhiyuan~Zhong,\\
Steven~C.H.~Hoi,~\IEEEmembership{Senior Member,~IEEE,}
Chun~Chen,~\IEEEmembership{Member,~IEEE,}%
\IEEEcompsocitemizethanks{\IEEEcompsocthanksitem Qingqun Ning, Jianke Zhu, Zhiyuan Zhong and Chun Chen are with the College of Computer Science, Zhejiang University, Hangzhou, China, 310027.\protect\\
E-mail: \{ningqingqun,jkzhu,zyzhong,chenc\}@zju.edu.cn.
\IEEEcompsocthanksitem Steven C.H. Hoi is with School of Information Systems, Singapore Management University, Singapore.\protect\\
E-mail: chhoi@smu.edu.sg
\IEEEcompsocthanksitem Jianke Zhu is the Corresponding Author.}
\thanks{}}
\maketitle

\begin{abstract}
    Fast Approximate Nearest Neighbor~(ANN) search technique for high-dimensional feature indexing and retrieval is the crux of large-scale image retrieval. A recent promising technique is Product Quantization, which attempts to index high-dimensional image features by decomposing the feature space into a Cartesian product of low dimensional subspaces and quantizing each of them separately. Despite the promising results reported, their quantization approach follows the typical hard assignment of traditional quantization methods, which may result in large quantization errors and thus inferior search performance. Unlike the existing approaches, in this paper, we propose a novel approach called Sparse Product Quantization~(SPQ) to encoding the high-dimensional feature vectors into sparse representation. We optimize the sparse representations of the feature vectors by minimizing their quantization errors, making the resulting representation is essentially close to the original data in practice. Experiments show that the proposed SPQ technique  is not only able to compress data, but also an effective encoding technique. We obtain state-of-the-art results for ANN search on four public image datasets and the promising results of content-based image retrieval further validate the efficacy  of our proposed method.
\end{abstract}

\begin{IEEEkeywords}
Approximate Nearest Neighbor Search, Sparse Representation, Product Quantization, Image Retrieval 
\end{IEEEkeywords}

%
\IEEEpeerreviewmaketitle

\section{Introduction}


\IEEEPARstart{I}{mage} retrieval is an important technique for many multimedia applications, such as face retrieval~\cite{kafai2014discrete}, object retrieval~\cite{chum2007total}, and landmark identification~\cite{landmark}. For large-scale image retrieval tasks, one of the key components is an effective indexing method for similarity search~\cite{wu2014sparse, cai2014scalable}, particularly on high-dimensional feature space~\cite{Lowe:04, zhang2014usb, rublee2011orb, spyromitros2014comprehensive}. 
Similarity search, \textit{a.k.a.}, nearest neighbor~(NN) search, is a fundamental problem. Due to the curse of dimensionality, exact NN search for high-dimensional data is extremely challenging and expensive. To overcome the issue, extensive research efforts have been devoted to approximate nearest neighbor~(ANN) search methods, such as hashing~\cite{weiss2009spectral,li2013spectral,LSH}, tree-based methods~\cite{silpa2008optimised,muja_flann_2009,flann_pami_2014}, and vector quantization~\cite{gray1984vector, JDS11}, which attempt to find the nearest neighbor with high probability using much less searching time and memory cost. 

In this paper, we focus on developing a vector quantization~(VQ) method for similarity search, which is a typical approach to effectively encoding the data for ANN search. A codebook is learnt  and every feature vector in the database can be represented by one of the most similar vectors in the codebook, typically named as ``codeword". Then VQ directly employs the index of the codeword to represent the original data vector, which typically has only a few bits. In addition, the similarity between query and the data vector in database can be approximated by calculating the distance between query and codebook vector. This greatly reduces the computational cost and searching time.




\begin{figure}[t]
    \begin{center}
        \includegraphics[width=0.95\columnwidth]{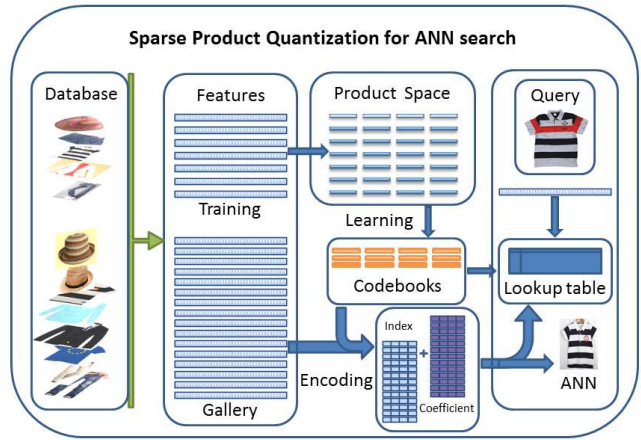}
        \caption{The system view of Sparse Product Quantization for ANN search.}
        \label{overview}
    \end{center}
\end{figure}

In general, VQ requires more bits in order to reduce quantization distortion. Since the size of codebook increases exponentially with respect to the total number of encoded bits, VQ-based method is ineffective for the data with high dimensionality. To tackle this issue, Product Quantization (PQ)~\cite{JDS11} has recently been shown a promising paradigm for efficiently indexing the high-dimensional image features. Different from other Hashing-based methods, it decomposes the high-dimensional space into a Cartesian product of low dimensional subspaces and quantize each of them separately. Since the dimensionality of each subspace is relatively small, using a small-sized codebook is sufficient to obtain the satisfied searching performance.

Although computational cost can be effectively reduced by diving the long vector into small segments, PQ may fail to retrieve the exact nearest neighbor of a query with high probability due to the high quantization distortion. As discussed in~\cite{PAMI_GeHK}, this will eventually yield lower search accuracy compared to VQ. To deal with this problem, several remedies have recently been proposed. Gong and Lazebnik~\cite{ITQ} presented an iterative quantization approach which maps data onto binary codes for fast retrieval. Cartesian K-means~\cite{ckmeans} and Optimized Product Quantization~\cite{GeHK013} share the same idea of rotating the original data to minimize the quantization error. These methods including PQ essentially follow the same framework of vector quantization, which all suffer from the inevitable nontrivial quantization distortion.







To address the above limitations, in this paper, we propose a novel approach called Sparse Product Quantization (SPQ) to encoding the high-dimensional vector of image features, where the sparse coding technique is introduced into approximate nearest neighbor search. Motivated by soft assignment~\cite{DBLP:conf/cvpr/PhilbinCISZ08}, we intend to find the sparse representation for each segment of feature vector rather than hard assignment used in PQ.  Specifically, a feature vector is decomposed into Cartesian product of the low dimensional subspaces, where the short vector in each subspace is approximated by the linear combination of several vectors from the codebook. Fig.~\ref{overview} illustrates the overview of SPQ for ANN search. 
We formulate the encoding stage as a sparse optimization problem and solve it by employing a popular greedy algorithm. The Euclidean distance between two vectors can be efficiently estimated from their sparse product quantization through simple table lookups. Moreover, the proposed method is able to take advantage of the very efficient SSE implementation using SIMD instructions, which can greatly reduce the computational overhead. Thus, the computational time of our presented method is comparable to the PQ method's while the precision of SPQ outperforms that of PQ at a very large margin. 
In contrast to the computationally intensive clustering algorithm used in all the VQ-based paradigms, we employ the sparse structure along with the fast stochastic online algorithm~\cite{Mairal_2009_ODL,Mairal_2010_OLM} to efficiently generate the codebook, which optimizes the sparse representation of data vectors according to their quantization errors. Consequently, the proposed representation is essentially close to the original data in practice even with a few basics. The empirical evaluation demonstrates that the presented method yields state-of-the-art ANN search results and outperforms the popular approaches on the application of image retrieval. 

The rest of this paper is organized as follows. Section 2 reviews related work. Section 3 introduces basics of VQ and propose our sparse vector quantization. Section 4 presents the proposed sparse product quantization for ANN search. Section 5 discusses our experimental results in detail and finally Section 6 concludes this work.





\section{Related Work}

Fast NN search is a fundamental research topic which is extensively studied in literature such as multimedia application, image classification, and machine learning. Our work is related to approximate NN search methods, which can be roughly grouped into three categories: Hashing-based methods~\cite{weiss2009spectral,Transform_coding,conf/mm/WangWYL13}, KD-tree~\cite{Bentley:1975:MBS:361002.361007}, and Vector Quantization work~\cite{JDS11,GeHK013}. 

Hashing-based ANN search approach has received lots of attention. Most of them employ either random projection or the learning-based methods to generate compact binary codes. 
As a consequence, the similarity between two data vectors is approximately represented by the Hamming distance of their hashed codes. Random projection is an effective approach which preserves pairwise distances for data points. The most representative example is Locality Sensitive Hashing~(LSH)~\cite{LSH,Shakhnarovich:2006:NML:1197919}. 
According to the Jonson Lindenstrauss Theorem~\cite{johnson84extensionslipschitz}, LSH needs $\mathcal{O}(\ln \ n/\epsilon^2)$ random projections to preserve the pairwise distances, where $\epsilon$ is the relative error. Hence, LSH needs to employ the code with long bit length in order to boost the projection performance, which leads to both high computational cost and huge storage requirement. 
On the other hand, learning-based hashing methods~\cite{weiss2009spectral,Transform_coding,conf/mm/WangWYL13} try to learn the structure of input data. Most of these algorithms generate the binary codes by employing the spectral properties of the data affinity matrix, i.e., item-item similarity. Some other hashing methods also employ multi-modal data~\cite{wu2014sparse} or semantic information~\cite{li2013spectral}. Despite achieving promising gain with relatively short codes, these methods often fail to make significant improvement as code length increases~\cite{conf/cvpr/JolyB11}.

	
\begin{figure*}[htbp]
    \centering
    \subfloat[]{\includegraphics[width=0.8\columnwidth]{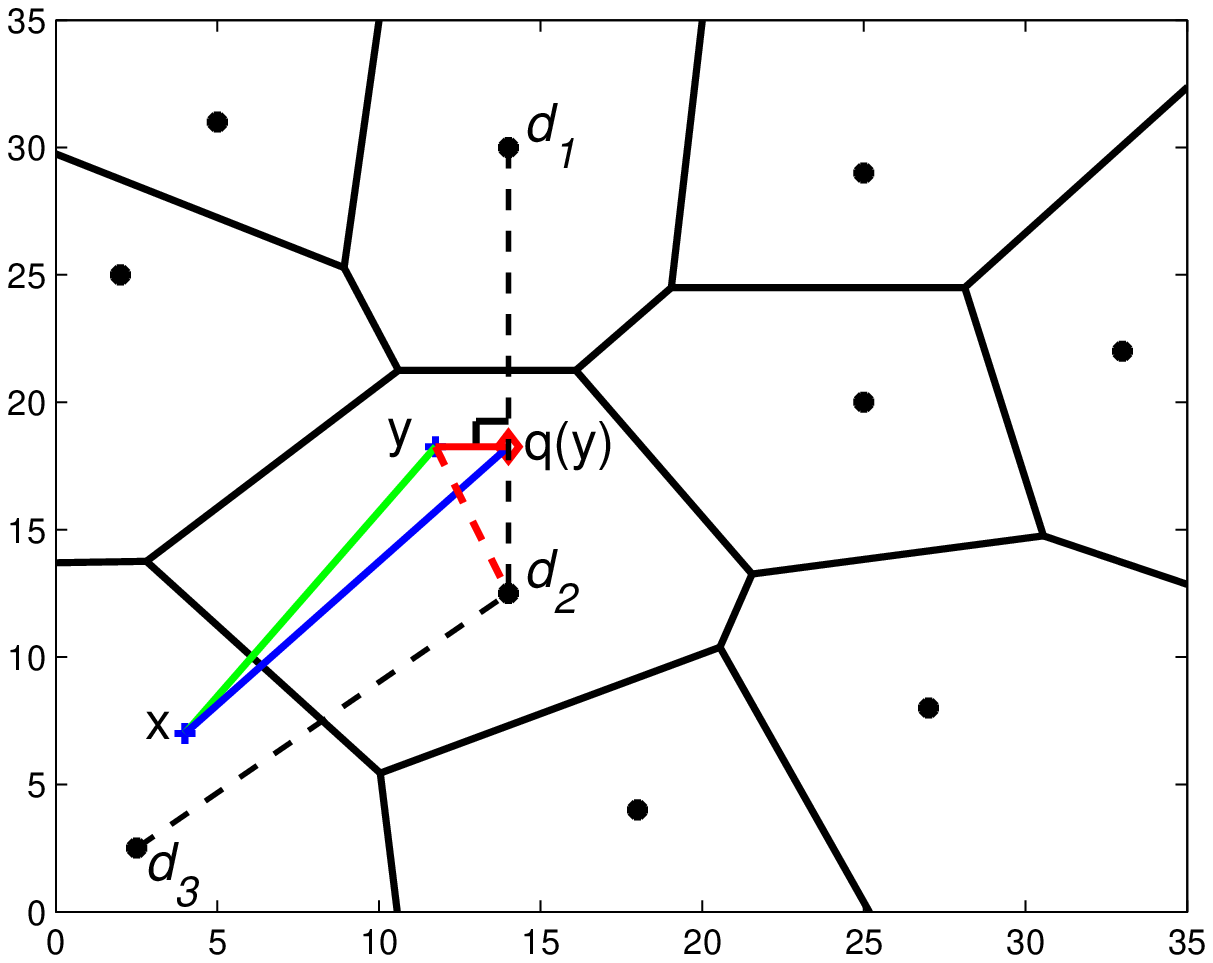}\label{fig:adc}}%
    \subfloat[]{\includegraphics[width=0.8\columnwidth]{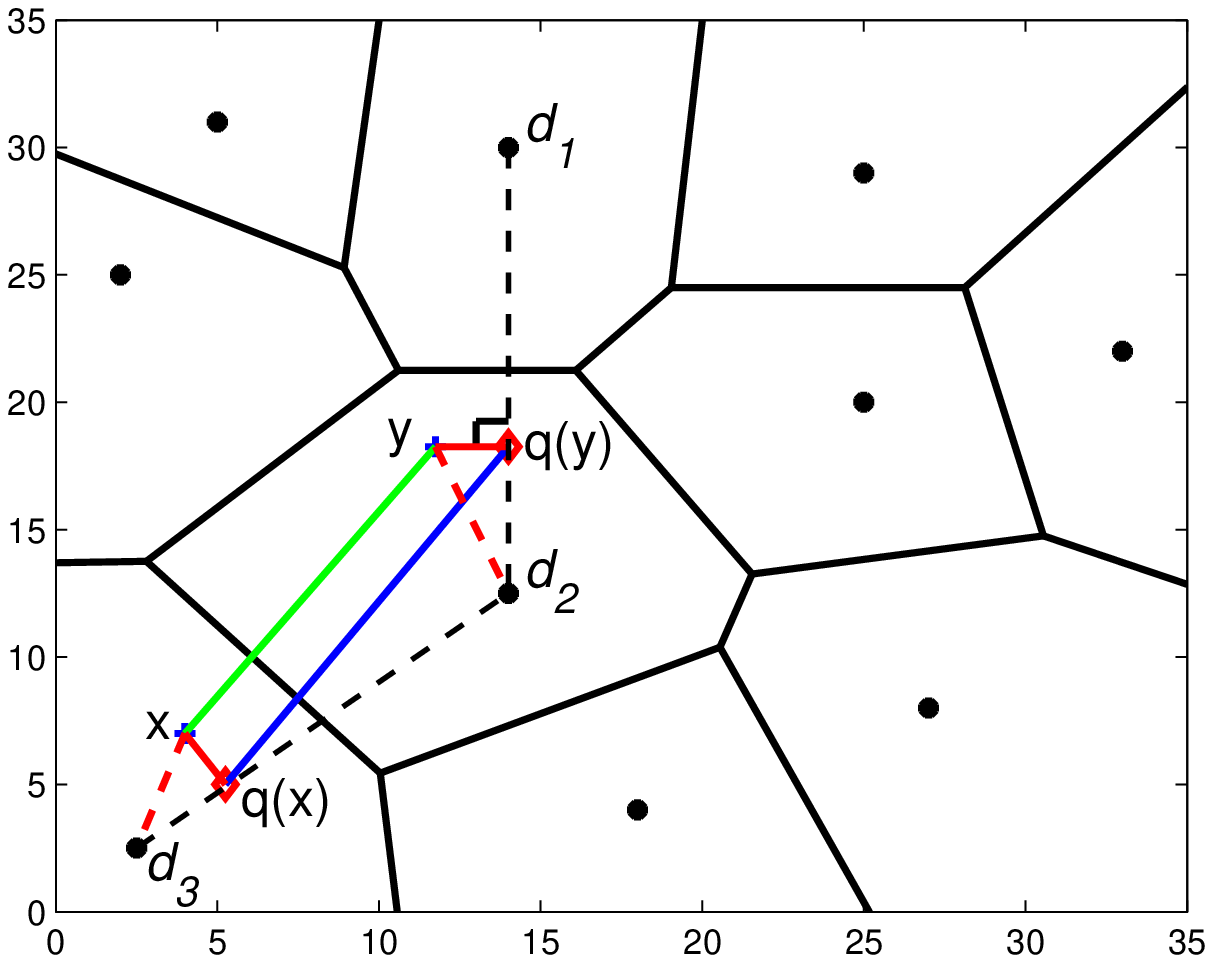}\label{fig:sdc}}%
    \caption{2D toy example of sparse vector quantization. $\mathbf{x}$ and $\mathbf{y}$ denote query and vector in gallery. $q(\mathbf{x})$ and $q(\mathbf{y})$ represent their quantization vectors. Instead of using hard assignment by the nearest center, we employ the sparse representation of the codebook with few word. Thus, $q(\mathbf{y})$ is the projection of $\mathbf{y}$ on line spanned by $\mathbf{d}_1$ and $\mathbf{d}_2$. As $d(\mathbf{y},q(\mathbf{y})) \leq d(\mathbf{y},\mathbf{d}_2)$, the distortion of our method is always smaller than VQ's. \protect \subref{fig:adc} Asymmetric Distance Computation (ADC); and \protect \subref{fig:sdc} Symmetric Distance Computation (SDC).}
    \label{adcsdc}%
\end{figure*}

The second group of research aims at speeding up the ANN search with KD-tree~\cite{Bentley:1975:MBS:361002.361007}. The expected complexity of KD-tree search is $\mathcal{O}(D\, \log\,n)$, while the brute-force search is $\mathcal{O}(nD)$. 
Unfortunately, for high dimension data KD-tree are not much more efficient than the brute-force exhaustive search~\cite{Weber:1998:QAP:645924.671192} due to the curse of dimensionality. Nevertheless, both randomized KD-trees~\cite{Optimised_kdtree,Randomized_kdtree} and hierarchical K-means~\cite{Nister:2006:SRV:1153171.1153548} improve the performance of KD-tree. 
In particular, these two methods are included in FLANN~\cite{muja_flann_2009, flann_pami_2014}, which automatically selects the best algorithm and optimal parameters depending on the dataset. FLANN is much faster than other publicly available ANN search software. However, KD-tree approaches need fully access to the data and thus cost much more memory in searching stage.

The third group of related work is about Vector Quantization based approaches, which try to approximate data vectors with codewords in the codebook. J\'egou et al.~\cite{JDS11} proposed an efficient product quantization (PQ) recently. 
The key of PQ is to decompose the feature space into a Cartesian product of low dimensional subspaces and quantize each one separately using their corresponding predefined codebook. Then, the distance between the query and a vector in gallery set can be computed by either symmetric distance computation (SDC) or asymmetric distance computation (ADC). Also, the inverted file system is employed to conduct non-exhaustive search efficiently. 
Empirically, PQ has been shown to significantly outperform various hashing-based methods in terms of accuracy. As discussed in~\cite{JDS11}, the prior knowledge on the underlying structures of input data is essential to VQ. 
Most recently, Ge et. al~\cite{GeHK013} consider PQ as an optimization problem that minimizes the quantization distortions by searching for the optimal codebooks and space decomposition. Due to the inherent nature of VQ~\cite{gray1984vector}, it is hard for these methods to evaluate the impact of quantization error on the ANN search performance. We should mention that a work called Product Sparse Coding~\cite{ge2014product} was published recently. However, it substantially differs from our work as it brings a strategy for sparse coding, though we both have relationship with product method and sparse coding.


Finally, our work is closely related to soft-assignment~\cite{DBLP:conf/cvpr/PhilbinCISZ08}, which has been introduced into the context of object retrieval~\cite{Philbin07} in order to reduce the quantization error. The key idea of soft-assignment is to map the original high-dimensional descriptor to a weighted combination of multiple visual words rather than hard-assigned onto a single word as in previous work~\cite{Sivic03,Philbin07}. Still, this representation is just incorporated into a standard tf-idf architecture. Despite requiring extra storage and computational cost, soft-assignment always results in lower quantization distortion and thus yields a significant improvement of retrieval performance in practice.

\section{Sparse Vector Quantization}
In this section, we first briefly review basics of Vector Quantization, and then introduce the proposed Sparse Vector Quantization (SVQ), followed by discussing the codebook training method for SVQ.

\subsection{Vector Quantization}
Vector Quantization (VQ)~\cite{gray1984vector} is a classical technique for data compression. It divides a dataset into some groups, where each vector is represented by the centroid of its corresponding group. More formally, given a vector $\mathbf{x}\in \mathbb{R}^D$, VQ maps $\mathbf{x}$ to the nearest codeword of a pre-trained codebook $\mathcal{C} = \{\mathbf{c}_i, i=1,2,...,k \}$ as follows:
\begin{equation}
    VQ: \mathbf{x} \rightarrow  q(\mathbf{x}) = \arg\min_{\mathbf{c} \in \mathcal{C}} d(\mathbf{x},\mathbf{c}).
\label{eq:vq}
\end{equation}
where $d(\cdot)$ is a distance metric. In particular, the distance used in this paper is Euclidean distance: $d^2(x,c)=\|x-c\|^2$. The encoding map $q(\cdot)$ is called quantizer which is the most important component of VQ. Therefore, the quantization distortion or reconstruction error $e(\mathbf{x})$ of $\mathbf{x}$ is defined as:
\begin{equation}
    e(\mathbf{x}) = d^2(\mathbf{x},q(\mathbf{x})) = \|\mathbf{x}-q(\mathbf{x})\|^2 = \min_{\mathbf{c}\in \mathbf{C} }\| \mathbf{x} - \mathbf{c}\|^2
\end{equation}
Given the codebook $\mathcal{C}$, the quantization of $\mathbf{x}$ is computed by solving the minimization problem in Eqn.~(\ref{eq:vq}). Typically, it can be simply represented by the Euclidean distance between the query and its corresponding codeword in $\mathcal{C}$.


In general, there are two kinds of ANN search methods according to different forms of queries. One is called \textit{Symmetric Distance Computation (SDC)}, in which both query and database vectors are quantized into codes. The other is called \textit{Asymmetric Distance Computation (ADC)}, where only the database vectors are quantized.

\subsection{Sparse Vector Quantization}

One key limitation of VQ is that it assigns the original vector to the single nearest codeword in the codebook. This hard assignment strategy
can lead to relatively large quantization distortion which limits the performance of VQ.

Motivated by the success of soft assignment~\cite{DBLP:conf/cvpr/PhilbinCISZ08}, instead of using the hard assignment as in VQ, we employ the sparse representation of multiple codewords to represent the original feature vector.

Fig.~\ref{adcsdc} shows a 2D-toy example to illustrate the key idea of our proposed method. Let $\mathbf{x}$ and $\mathbf{y}$ denote a query and the vector in gallery set respectively, and $q(\mathbf{x})$ and $q(\mathbf{y})$ represent the quantization vector for $\mathbf{x}$ and $\mathbf{y}$, respectively. VQ simply sets $q(\mathbf{y})$ to point $d_2$ by hard assignment, and similarly sets $q(\mathbf{x})$ to $d_3$. Thus, the quantization distortion for $\mathbf{y}$ is $d(\mathbf{y},\mathbf{d}_2)$. In this work, we employ the linear combination of two words $d_1$ and $d_2$ to represent $\mathbf{y}$. Therefore, $q(\mathbf{y})$ is the projection of $\mathbf{y}$ on the line spanned by $\mathbf{d}_1$ and $\mathbf{d}_2$. It is clear that the quantization distortion by VQ is always larger than that of the sparse quantization, since $d(\mathbf{y},q(\mathbf{y})) \leq d(\mathbf{y},\mathbf{d}_2)$.

As $q(\mathbf{y})$ lies on the line ($\mathbf{d}_1,\mathbf{d}_2$), we assume $ \mathbf{y} \approx q(\mathbf{y})= \alpha \mathbf{d}_1+ \beta \mathbf{d}_2$. Note that the coefficients $\alpha$ and $\beta$ can be easily computed by solving the linear equation. As illustrated in Fig.~\ref{adcsdc}, we can compute the distance $d(\mathbf{x},\mathbf{y})$ as follows:
\begin{equation}
\begin{split}
       d^2(\mathbf{x},\mathbf{y}) &= \| \mathbf{x}-\mathbf{y} \|^2
              \\& =\| \mathbf{x}\|^2 + \| \mathbf{y}\|^2 - 2 \langle \mathbf{x}, \mathbf{y} \rangle
              \\& \approx \| \mathbf{x}\|^2 + \| \mathbf{y}\|^2  - 2\langle \mathbf{x}, q(\mathbf{y}) \rangle
              \\& = \| \mathbf{x}\|^2 + \| \mathbf{y}\|^2 + 2\langle \mathbf{x},\alpha \mathbf{d}_1+ \beta \mathbf{d}_2 \rangle
              \\& =\| \mathbf{x}\|^2 + \| \mathbf{y}\|^2  + 2 (\langle \alpha \mathbf{x},\mathbf{d}_1 \rangle + \langle\beta \mathbf{x},\mathbf{d}_2 \rangle)
\label{egadc}
\end{split}
\end{equation}
where $\langle \cdot,\cdot \rangle $ denotes the dot product.
The above equation calculates the ADC distance. Also, we can calculate SDC distance $d(q(\mathbf{x}),q(\mathbf{y}))$ using the similar approximation method.



Before introducing SVQ, we first give an equivalent formulation of VQ. We stack the codebook $\mathcal{C}$ into a $D \times k$ matrix $C$, in which each of its columns is a word. Let $k$ denote the size of codebook $\mathcal{C}$, we can rewrite Eqn.~(\ref{eq:vq}) as the following optimization problem:
\begin{equation}
    \frac{1}{2}\min_{\alpha}\| C {\alpha} - \mathbf{x}\|^2,  \quad s.t.\;  \|{\alpha}\|_0 = 1 ,\alpha \in \{0,1\}^k
\label{eq:vq2}
\end{equation}
${\alpha}$ is a $k$ dimensional column vector, in which the value of each element is either zero or one. Obviously, the above optimization in Eqn.~(\ref{eq:vq2}) is equivalent to hard assignment by imposing very strict constraints on variable $\alpha$ to choose the nearest word from matrix $C$ given the input vector $\mathbf{x}$.

As in the above discussion, it can be observed that searching accuracy for ANN is directly related to the bound of Eqn.~(\ref{eq:vq2}) rather than its solution. To this end, we relax the constraints in Eqn.~(\ref{eq:vq2}) so as to obtain a lower bound. This will implicitly yield better ANN searching performance. Specifically, we relax the constraint in Eqn.~(\ref{eq:vq2}) as follows:
\begin{equation}
   \frac{1}{2} \min_{\alpha}\| C {\alpha} - \mathbf{x}\|^2,  \quad s.t.\;  \|{\alpha}\|_0 \le L ,\alpha \in \mathbb{R}^k
    \label{eq:spq}
\end{equation}
where $L$, named as \textit{sparse level}, denotes the number of codewords selected to encode $\mathbf{x}$. It can be seen that such relaxation not only increases $L_0$ norm of sparse representation but also expands the space of $\alpha$. Obviously, Eqn.~(\ref{eq:vq2}) can be viewed as a special case of Eqn.~(\ref{eq:spq}) when $L=1$. Therefore, we can obtain a lower bound for quantization distortion. Intuitively, the above formulation employs the linear combination of $L$ words in codebook rather than using only single word as VQ to approximate the original input vector. Our empirical study shows that using just two words is sufficient to yield significant gain over the hard assignment.


Eqn.~(\ref{eq:spq}) is well-known as an NP-Hard problem. To tackle this issue, we take advantage of an effective greedy algorithm called Orthogonal Matching Pursuit (OMP)~\cite{Mairal_2009_ODL,Mairal_2010_OLM}.
OMP updated all the extracted coefficients by computing the orthogonal projection of the vector residual onto the set of codewords selected so far.
As $L$ is usually set to two, there are at most two non-zero elements in the coefficient vector ${\alpha}$. As the sparse property of the representation is essential to fast NN search, we thus name our method as Sparse Vector Quantization (SVQ).

\subsection{Codebook Training}

Remember that we assume the codebook of each approach has been given in previous analysis. In this part, we will show how to obtain the codebook.

The first common and straightforward method is to find the codebook by directly minimizing the quantization error on the training set $\{x| x\in X\}$. In the case of VQ, the codebook is obtained by solving Eqn.~(\ref{eq:vq}) or Eqn.~(\ref{eq:vq}) on $X$, and this is equivalent to running an iterative k-means clustering algorithm where the centroids of the resulting $k$ clusters are treated as the codebook.

For SVQ, minimizing the quantization error is equal to the following problem:
\begin{equation}
    \min_{C \in \mathbb{R}^{D\times k} ,\alpha_i \in \mathbb{R}^k } \frac{1}{2}\sum_{i=1}^n \| C {\alpha_i} - \mathbf{x}_i\|^2 , \quad s.t.\;  \|{\alpha}\|_0 \le L.
    \label{eq:codebooksvq}
\end{equation}
It is NP-hard.
We can alternate between $C$ and $\alpha$ to solve this problem. When $C$ is fixed, we have shown how to solve it in previous section. Notice that here what we care is the codebook $C$. Then we can further relax the contraints by using an $L_1$-norm constraint which can also yield sparse solutions. In section~\ref{sec:setting}, we will see that both methods is applicable to our method. When $\alpha$ is fixed, it becomes an uncontrained least square problem.
In our implementation we employ the stochastic/online optimization algorithm~\cite{Mairal_2009_ODL,Mairal_2010_OLM} to solve the above optimization problem for learning the codebook, where the learned codebook can be excellently fitted for the sparse coding tasks. Since the algorithm is based on stochastic optimization, it is even faster than conventional k-means clustering method.




In general, VQ-based methods heavily rely on a good codebook, which is important to reduce the quantization distortion.
Due to its intrinsic limitedness, k-means is often difficult to generate a good one.  In the experiment, we will show that, 
In contrast to other VQ-based methods such as PQ and OPQ, our proposed SPQ method is not limited to any specific codebook learning method. 


\section{Sparse Product Quantization}

To facilitate the practical ANN search, we propose an efficient Sparse Product Quantization approach by extending the product quantization with the proposed SVQ technique in order to further reduce the computational overhead.

\subsection{Product Quantization}
Following the idea of Product Quantization (PQ)~\cite{JDS11}, we decompose the high-dimensional space into a Cartesian product of low dimensional subspaces and then perform sparse vector quantization in each subspace separately. Specifically, a vector $\mathbf{x}$ is viewed as the concatenation of $m$ subvectors:
$\mathbf{x} = [\mathbf{x}^1,\mathbf{x}^2,...,\mathbf{x}^m]$ and the codebook is defined as: $\mathcal{C}=\mathcal{C}^1\times \mathcal{C}^2\times...\times \mathcal{C}^m$.

For PQ, each subvector is mapped onto a sub-codeword from its corresponding codebook:
\begin{equation}
\begin{split}
    \mathbf{x} & = [\mathbf{x}^1,\mathbf{x}^2,...,\mathbf{x}^m].\\
    PQ: \mathbf{x^i} & \rightarrow  q_i(\mathbf{x}) = \arg\min_{\mathbf{c} \in \mathcal{C}^i} d(\mathbf{x^i},\mathbf{c}), \\ i &=1,2,...,m.
\end{split}
\label{eq:pq}
\end{equation}
where $q_i$ is a quantizer for the $i$-th subvector of $\mathbf{x}$. Practically, $\mathbf{x}$ is equally partitioned so that all subvectors $\mathbf{x}^i\in \mathbb{R}^{D/m}$ and $D$ is a multiple of $m$.
Note that each subvector is encoded according to the different codebook. In this case, any word $\mathbf{c}$ of $\mathbf{x}$ in codebook $\mathcal{C}$ will be the concatenation of $m$ sub-codewords: $\mathbf{c} = [\mathbf{c}^1,\mathbf{c}^2,\mathbf{c}^i,...,\mathbf{c}^m]$, with each $\mathbf{c}^i \in \mathcal{C}^i,\ i\in [1,m]$.

Let $Q(\mathbf{x}) = [ q_1(\mathbf{\mathbf{x}}), q_2(\mathbf{x}),\allowbreak ...,q_m(\mathbf{x})]$ denote the PQ of $\mathbf{x}$. Then, the quantization distortion of $\mathbf{x}$ by PQ is defined as follows:
 \begin{equation}
     e(\mathbf{x}) = d^2(\mathbf{\mathbf{x}},Q(\mathbf{\mathbf{x}})) = \sum_{i=1}^m\|\mathbf{x}^i-q_i(\mathbf{x})\|^2
\end{equation}

Usually, we need to quantize a set of vectors $ \mathcal{X} = \{\mathbf{x}_i, i=1,...,n\}$ rather than single one. Hence, the quantization distortion of $\mathcal{X} $ is defined as $E(\mathbf{X}) = \sum_{\mathbf{x}\in \mathcal{X}} e(\mathbf{x})$.

As in Eqn.~(\ref{eq:pq}), it can be easily observed that PQ divides Eqn.~(\ref{eq:vq}) into $m$ sub-VQ problems and therefore addresses it separately. Therefore, PQ method enjoys the merit of providing the compact coding scheme for high-dimensional data while yielding accurate result for fast approximate nearest neighbor search. However, the unavoidable quantization error limits its performance of searching accuracy due to the inherent nature of vector quantization~\cite{gray1984vector}.

Intuitively,  better reconstruction that means having lower quantization distortion indicates  better search accuracy .
In next section, we will introduce an approach which can effectively reduce the quantization distortion.


\subsection{Sparse Product Quantization}

In the proposed sparse vector quantization, we represent each item $\mathbf{x}$ in the database as follows:
\begin{equation}
    \mathbf{x} \approx C \cdot {\alpha}.
\end{equation}
where ${\alpha}$ is a sparse vector with a few non-zero elements.

Motivated by product quantization, in this paper, we employ the proposed SPQ scheme with slight modification by replacing $\mathbf{x}$ with its subvector $\mathbf{x}^i$. Therefore, we can approximate $\mathbf{x}^i$ through the following equation:
\begin{equation}
    \mathbf{x}^i \approx C^i {\alpha}^i,\ i=1, 2, ..., m,
    \label{appro}
\end{equation}

We can prove that the quantization distortion of SPQ is upper bounded by that of PQ.
Remember that SVQ is a relaxation version of VQ. Thus, the bound for quantization distortion of SVQ is lower than that of VQ. 
In the case of PQ and SPQ, their distortions are the sum of distortions for each subvector. With respect to each subvector, the situation is equal to that of VQ and SVQ.  Thus, we can conclude that the quantization distortion of SPQ is less than or equal to PQ's.

\renewcommand{\algorithmicrequire}{ \textbf{Input:}}
\renewcommand{\algorithmicensure}{ \textbf{Output:}}
\begin{algorithm}
    \caption{ANN Search with Sparse Product Quantization}
    \label{alg:ann}
    \begin{algorithmic}[1]
        \REQUIRE ~~\\  
        The database $X = [\mathbf{x}_1, \cdots, \mathbf{x}_n]$, the codebook size $k$, the subspace number $s$, the sparse level $L$, the query set $Q = [\mathbf{q}_1, \cdots, \mathbf{q}_m]$, the number of NN $p$.
        \ENSURE ~~\\
        The top $p$ ANN indexs $I$, the top $p$ ANN distances $D$
        
        \item[]
        \item[]
        $/*$ Encoding Stage $*/$
        \STATE Sample a subset $X_s$ of $X$.
        \FOR{each subspace $X_s^i$ of $X_s$}
        \STATE Using the fast stochastic online algorithm~\cite{Mairal_2009_ODL} to train a codebook $C^i$ of size $k$ on $X_s^i$ .
        \ENDFOR
        \FOR{each subspace $X^i$ of $X$}
    \STATE Compute sparse coefficient $A^i = \{\alpha_1^i, \cdots, \alpha_n^i\}$ on $C^i$ using Orthogonal Matching Pursuit algorithm, such that \begin{equation*} \alpha_j^i = \arg\min_\alpha\|C^i\alpha-\mathbf{x}_j^i\|^2, s.t.\; \|\alpha\|_0 \le L.\end{equation*}
        \ENDFOR
        
        \item[]
        \item[]
        $/*$ Query Stage $*/$
        \FOR{each query $\mathbf{q}$ of $Q$}
        \FOR{each subspace $\mathbf{q}^i$ of $\mathbf{q}$}
        \STATE Precompute lookup table $T^i$ with $\mathbf{q}^i$ and $C^i$.
                \STATE Using  $A^i$ and $T^i$ to compute the approximate distances $E^i$ to the database on this subspace.
            \ENDFOR
            \STATE Sum up the approximate distances $E = \sum_i E^i$.
            \STATE Search the top $p$ NNs based on $E$ and save them to $I$ and $D$.
        \ENDFOR

    \end{algorithmic}

\end{algorithm}

\subsection{Approximate Nearest Neighbor Search}
In the following, we discuss how to apply the proposed SPQ method to conduct ANN search towards large-scale image retrieval tasks.
The whole framework of our proposed SPQ approach of ADC version is summarized into Algorithm~\ref{alg:ann}.


In particular, to facilitate ANN search, we encode all the data vectors in the gallery using the proposed SPQ method.
Then, we compute the distance between a query $\mathbf{q}$ and the data in the gallery using two kinds of distance measures: ADC and SDC. 

According to the definition, ADC can be formulated as:
{\setlength\arraycolsep{2pt}
\begin{eqnarray}
d^2(\mathbf{q},\mathbf{x}) & = & \sum_{i=1}^m\|\mathbf{q}^i-\mathbf{x}^i)\|^2 \nonumber\\
& = & \sum_{i=1}^m  \|\mathbf{x}^i\|^2 + \|\mathbf{q}^i\|^2 - 2\langle \mathbf{x}^i, \mathbf{q}^i \rangle \nonumber\\
& = & \|\mathbf{x}\|^2 + \|\mathbf{q}\|^2  -2\sum_{i=1}^m   \langle C^i  {\alpha}^i ,\mathbf{q}^i \rangle.
\label{eq:adcdist}
\end{eqnarray}
To reduce the computational cost for ADC distance computation, we can either normalize $\mathbf{x}$ or precompute $\|\mathbf{x}\|^2$. Since ${\alpha}$ is an essentially sparse vector, it only requires several floating point operations to compute $\langle C^i  {\alpha}^i, \mathbf{q}^i\rangle $.

In the case of SDC computation, we employ sparse product quantization to approximate the query vector $\mathbf{q}$ as:
$\mathbf{q}^i \approx C^i \cdot \beta^i, i=1,2,...,m$  ,
Similarly, SDC is computed as:
{\setlength\arraycolsep{2pt}
\begin{eqnarray*}
d^2(\mathbf{q},\mathbf{x}) &=& \sum_{i=1}^m\|\mathbf{q}^i-\mathbf{x}^i)\|^2 \\
              & = & \sum_{i=1}^m \|\mathbf{x}^i\|^2 + \|\mathbf{q}^i\|^2 - 2\langle \mathbf{x}^i, \mathbf{q}^i\rangle\\
              & = & \|\mathbf{x}\|^2 + \|\mathbf{q}\|^2  -2\sum_{i=1}^m   \langle C^i  {\alpha}^i, C^i {\beta}^i\rangle
    \label{eq:sdcdist}
\end{eqnarray*}
For better illustration, Fig.~\ref{adcsdc} shows a 2D example of distance computation for both ADC and SDC.


\subsection{Complexity Analysis}

In the following, we give the detailed analysis on the complexity of our proposed SPQ scheme.

Let $D$ denote the dimensionality of each feature vector, $n$ denote the total number of items in the whole database, $m$ denote the number of subvectors in $\mathbf{x}$, and $k$ denote the size of each codebook. For a given query, it takes $\mathcal{O}(nD + kD +nmL)$ floating point multiplications to search its approximate nearest neighbor in database $\mathcal{X}$. Specifically, it requires $nD$ multiplications for the dot product of each database vector. This is required if the database is not normalized. Also, it takes $kD$ operations to compute the distance between query vector and vocabulary matrix $C$. We need $nmL$ multiplications to compute the dot product with the database vectors in sparse representation, which is the third term in Eqn.~(\ref{eq:adcdist}).

If all the database vectors have been normalized with unit $L_2$ norm offline, then $ \|\mathbf{x}\|^2 = \|\mathbf{q}\|^2  = 1 $. Therefore, the overall online time complexity to computing ADC distance can be reduced to $\mathcal{O}(kD+nmL)$. In the task of multimedia information retrieval, the dimensionality of each feature vector $D$ is far less than the total number of entries in database: $D \approx k \ll n$. Thus, the computational complexity of Eqn.~(\ref{eq:adcdist}) can be approximated to $\mathcal{O}(nmL)$. On the other hand, the complexity of brute-force NN search is $\mathcal{O}(nD)$. Thus, we can obtain substantial speedup using the proposed SPQ scheme. Moreover, our method is able to take advantage of the efficient SSE instructions to further reduce the multiplication computational time. Specifically, the searching time of our proposed SPQ is comparable to the original PQ while the precision of SPQ outperforms that of PQ at large margin. Additionally, the empirical study shows that SPQ is even faster than FLANN~\cite{flann_pami_2014} with the same recall rate.

Due to the inherent nature of soft-assignment, SPQ consumes more memory cost than hard-assignment methods inevitably. However, it is worth the memory because SPQ brings significant gain on precision improvement. 
According to the previous studies, FLANN is one of the most popular ANN search techniques that utilize tree structure. However, it fails to work for very large-scale datasets since it must load the whole dataset in memory when building the trees. By contrast, SPQ does not need to load the whole data in memory by employing the efficient inverted file structures, making it potentially more practical than FLANN for large-scale multimedia retrieval.





\section{Experiment}
In this section, we will first introduce our experimental testbed and the background of several state-of-the-art ANN methods we will compare with. Then we discuss the settings of our proposed method and furnish our results comparing with these methods. Finally, we show the application of our method on image retrieval.


\subsection{Experimental Testbed}
To examine the empirical efficacy of the proposed method, we conduct an extensive set of experiments for comprehensive performance evaluations on five datasets, including a synthetic dataset with Gaussian noises and four publicly available image feature collections. Each dataset is partitioned into three parts: training set, gallery set and query set. The details of these testbeds are summarized as follows: 1) SIFT dataset consists of one million local SIFT features~\cite{Lowe:04} with 128 dimensions, in which 100K samples are employed to learn the codebook. All the one million samples are treated as gallery set, and 10K samples are used for evaluation. Note that there is no overlap between the training set and the gallery set, since the former is extracted from Flickr images and the latter is from the INRIA Holidays images~\cite{jegou2008hamming}; 2) GIST~\cite{oliva2001modeling} is made of 960-dimensional global features. There are 50K samples used to learn the codebook. Similarly, one million samples in database are viewed as gallery set, and 1K samples are used for query evaluation. They are extracted from the tiny image set~\cite{torralba200880}, Holidays image set, and Holiday with Flickr1M set, respectively; 3) We perform empirical study on MNIST\footnote{\url{http://yann.lecun.com/exdb/mnist/}} as used in OPQ~\cite{PAMI_GeHK}, which is a 784-dimensional image set of hand-written digits with totally 70K images. In our experiment, we randomly sample 1K images as the queries and the remaining data are treated as the gallery set. To learn the codebook, we randomly pick 7K from the gallery set; 4) LabelMe dataset~\cite{russell2008labelme} contains 22,019 images, where each item is represented by a 512-dimensional GIST descriptor. Following~\cite{conf/mm/WangWYL13}, we randomly sample 2K images to form the query set and use the remaining data to form the gallery set; 5) We also synthesize a set of 128-dimensional vectors from independent Gaussian distributions. We choose 10K data to learn the codebook. 1M data is employed as gallery set, and 1K samples are used for query. All the compared methods are evaluated on the same dataset for each setting. To make it clear, Table~\ref{tb:dataset} summarizes the statistics of the datasets used in our experiments.


\begin{figure*}[t]
    \centering{
        \subfloat[]{
            \includegraphics[width=0.32\textwidth]{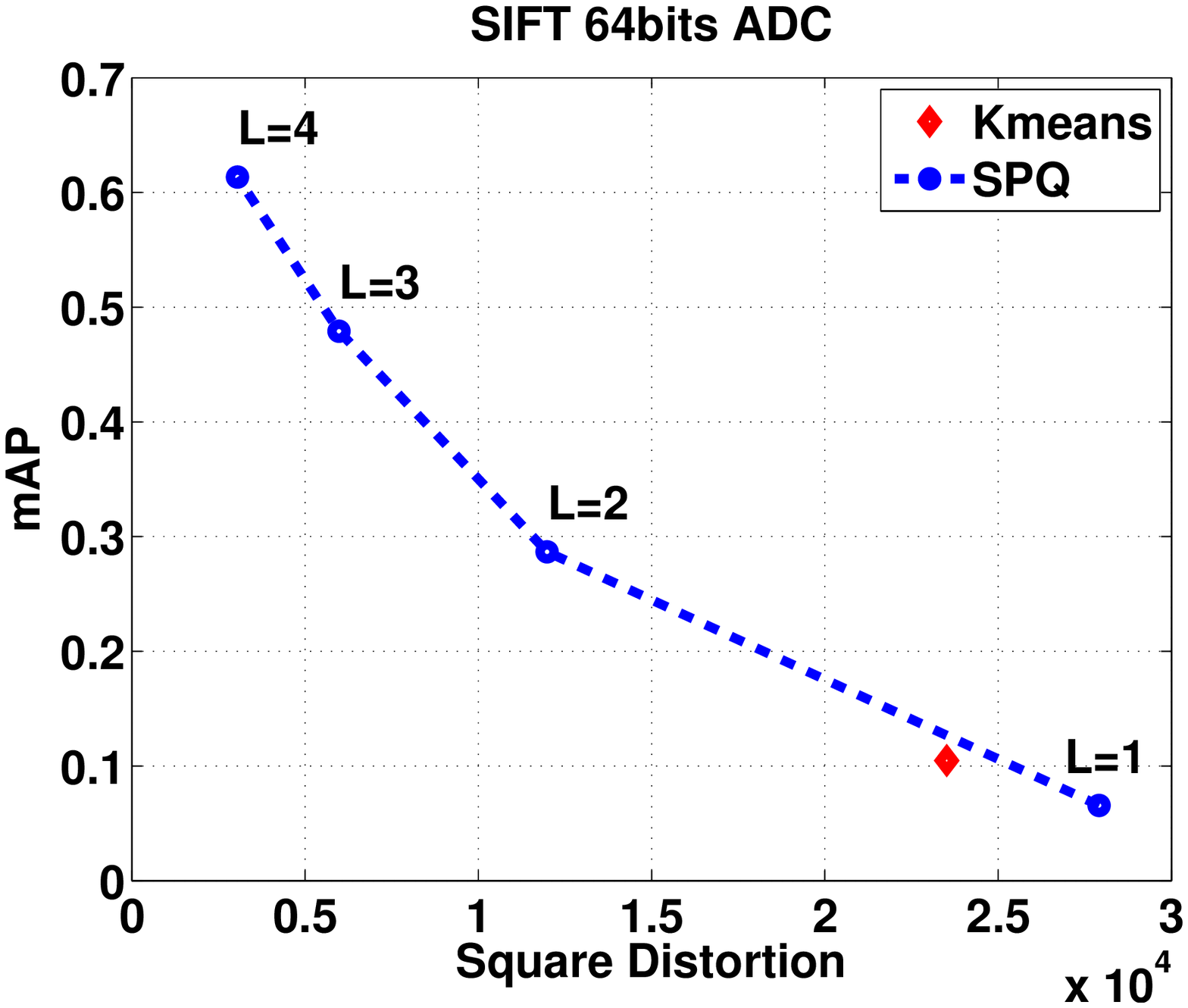}
        \label{subfig:sparselevel}}
        \subfloat[]{
            \includegraphics[width=0.32\textwidth]{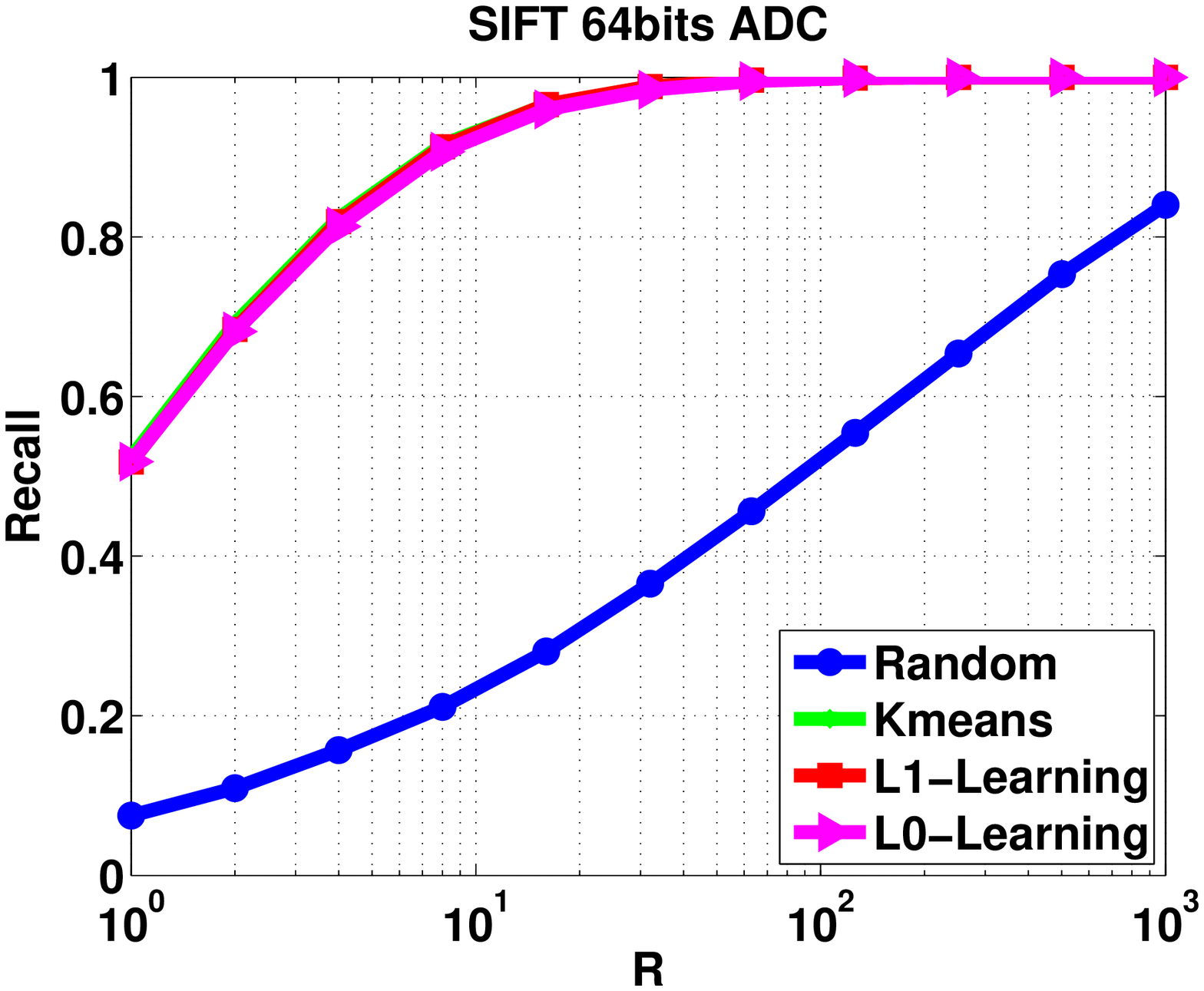}
        \label{subfig:codebook}}
        \subfloat[]{
            \includegraphics[width=0.32\textwidth]{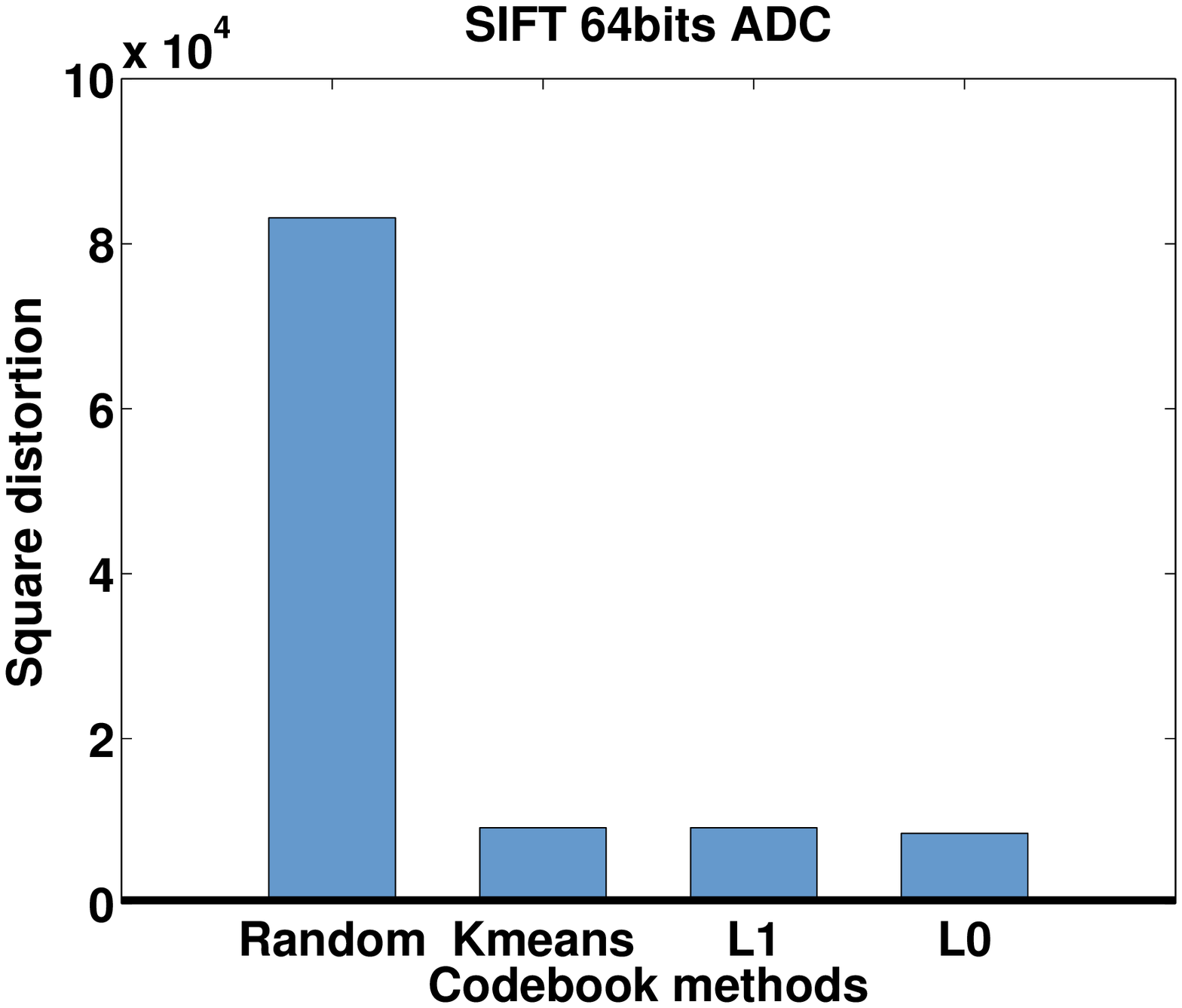}
        \label{subfig:codebookdist}}
    }
\caption{
    Setting experiments on SIFT dataset. 
    \protect \subref{subfig:sparselevel} mAP vs. Square Distortion under different sparse levels $L$. The square distortion decreases and mAP increases consistently when the sparse level $L$ increases.
    \protect \subref{subfig:codebook} Performance comparison on different codebook learning methods. Random denotes the random sampling method and L0-learning and L1-learning are the online dictionary learning algorithm with corresponded constraint~\cite{Mairal_2009_ODL}.  Our method perform very similar in real data with various codebook learning methods, except for random sampling method that contains no information of the data.
    \protect \subref{subfig:codebookdist} The quantization distortion on different codebook methods. It is easy to see the relationship between accuracy and distortion.
}
\label{fig:settings}
\end{figure*}

We compare our proposed Sparse Product Quantization (SPQ) approach with the following state-of-the-art methods.
\begin{itemize}
    \item Product Quantization~(PQ~\cite{JDS11}) tries to build codebook on Cartesian Product space, which is treated as baseline. IVFPQ refers to the PQ with the inverted file structure. All the results of PQ in the experiment are reproduced from the original implementation~\footnote{\url{http://people.rennes.inria.fr/Herve.Jegou/projects/ann.html}}.
    \item Optimized Product Quantization~(OPQ~\cite{GeHK013}) aims at finding an optimal space decomposition of PQ, which introduces two different solutions. Due to its superior performance, we only compare with the non-parametric solution using parametric one as a warm start. Similarly, we adopt their own implementation~\footnote{\url{http://research.microsoft.com/en-us/um/people/kahe/cvpr13/index.html}} with default settings.
    \item Cartesian K-means~(CK-means~\cite{ckmeans}) is yet another method to find the optimal space decomposition for PQ. 
    It is equivalent to OPQ while using the same initialization. The results of CK-means are produced from the publicly available implementation~\footnote{\url{https://github.com/norouzi/ckmeans/tree/}} with default setup.
    \item Iterative Quantization~(ITQ~\cite{ITQ}) is an effective binary embedding technique that can also be viewed as a vector quantization method. 
    \item Order Preserving Hashing~(OPH~\cite{conf/mm/WangWYL13}) is a state-of-the-art hashing method that learns similarity-preserving hashing functions.
    \item FLANN~\cite{flann_pami_2014} is the most popular open-source ANN search toolbox based on the framework of searching tree. It is able to automatically select the best algorithm and parameters for a given dataset.
\end{itemize}
The above methods can be roughly categorized into three groups: (i) VQ-based methods, including PQ, OPQ, and CK-means; (ii) hashing-based methods, including ITQ and OPH; and finally (iii) FLANN that is a searching tree-based method. In the following, we make the comparisons for each group separately.

\begin{table}[t]
    \centering
    \caption{Summary of our experimental testbeds}
    \label{tb:dataset}
	\begin{scriptsize}
    \begin{tabular}{|l|rrrrr|}
        \Xhline{2.5\arrayrulewidth}
        Dataset & SIFT & GIST & Random & MNIST & LabelMe\\
        \Xhline{2.5\arrayrulewidth}
        $d$ & 128 & 960 & 128 & 784 & 512\\
        \hline
        $Training$ & 100K & 50K & 10K & 10K & 10K\\
        \hline
        $Gallery$ & 1M& 1M & 1M & 60K& 20,019 \\
        \hline
        $Query$ & 10K & 1K & 10K & 1K & 2K\\
        \Xhline{2.5\arrayrulewidth}
    \end{tabular}
	\end{scriptsize}
\end{table}

In our empirical study, distortion is employed to measure the reconstruction performance for vector quantization. To evaluate the efficacy of ANN search methods, we employ the conventional performance metrics for multimedia information retrieval, including precision, recall and mAP. Precision means the average proportion of true NNs ranked first in the returned candidates, and recall denotes the proportion of true NNs of all queries is ranked. Moreover, mAP is the mean of Average precision over all the queries, which indicates the overall performance. All of our experiments were carried out on a PC with Intel Core i7-3770 3.4GHz processor and 16GB RAM using single thread.


\subsection{Settings}
\label{sec:setting}

We discuss the experimental settings for the proposed SPQ approach in the following.

\textbf{Sparse Level $L$} denotes the number of words to encode a feature vector by our method, which is critical to our method. Fig.~\ref{subfig:sparselevel} shows mAP  with respect to the quantization square distortions under different sparse levels on the SIFT dataset. Clearly, the square distortion decreases consistently when the sparse level $L$ increases, and at the same time mAP increases. Moreover, we found that the distortion drops significantly from level one to level two. Since the computational time and memory consumption grow with the sparse level, we set $L$ to 2 in the following experiments as a tradeoff between efficiency and accuracy. In section~\ref{sec:comp}, we will see that this sparse level is good enough to outperform the state-of-the-art methods.

\begin{figure*}[tbp]
    \centering
    \subfloat[]{
    \begin{minipage}{0.24\textwidth}
        \includegraphics[width=\textwidth]{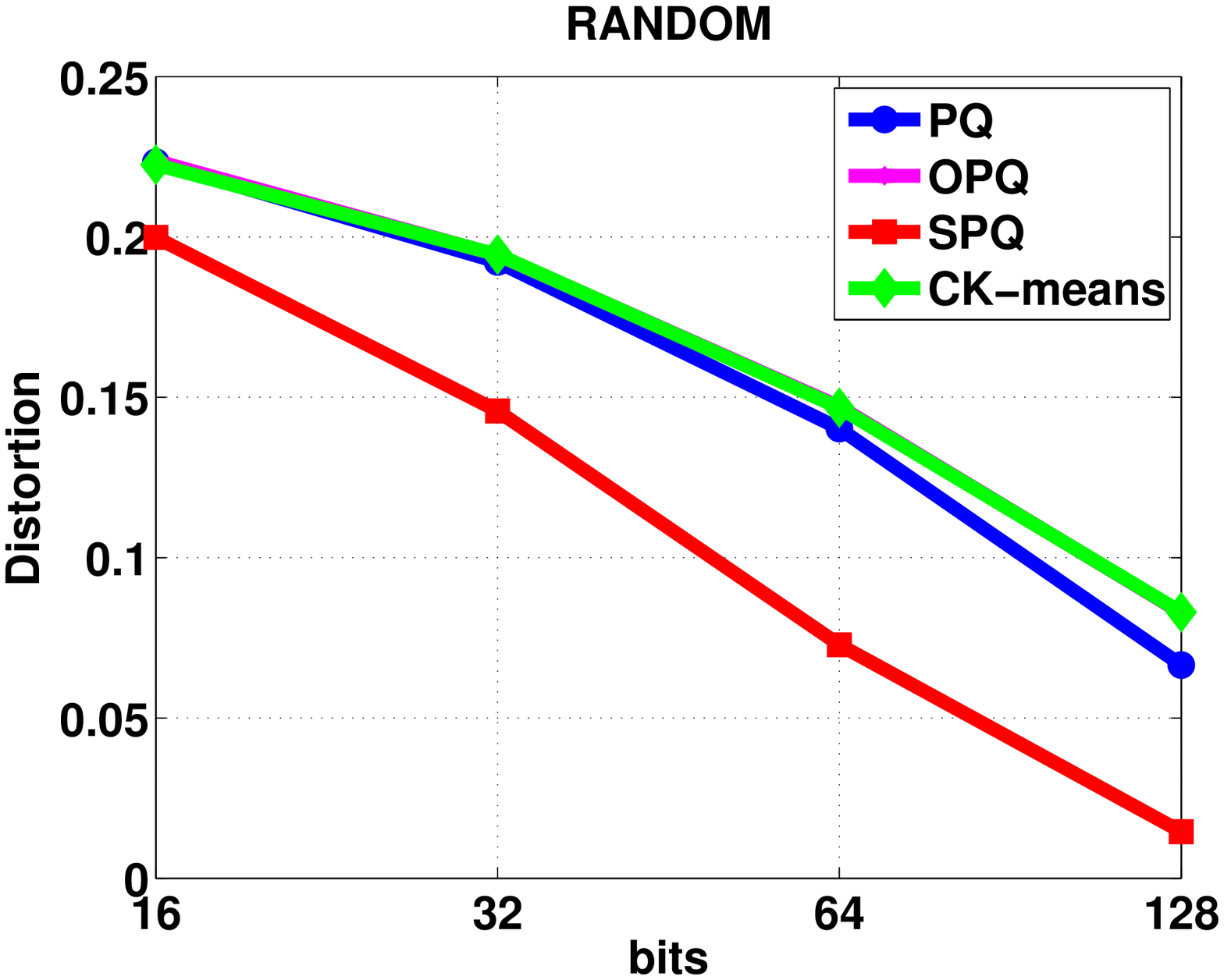}
        \includegraphics[width=\textwidth]{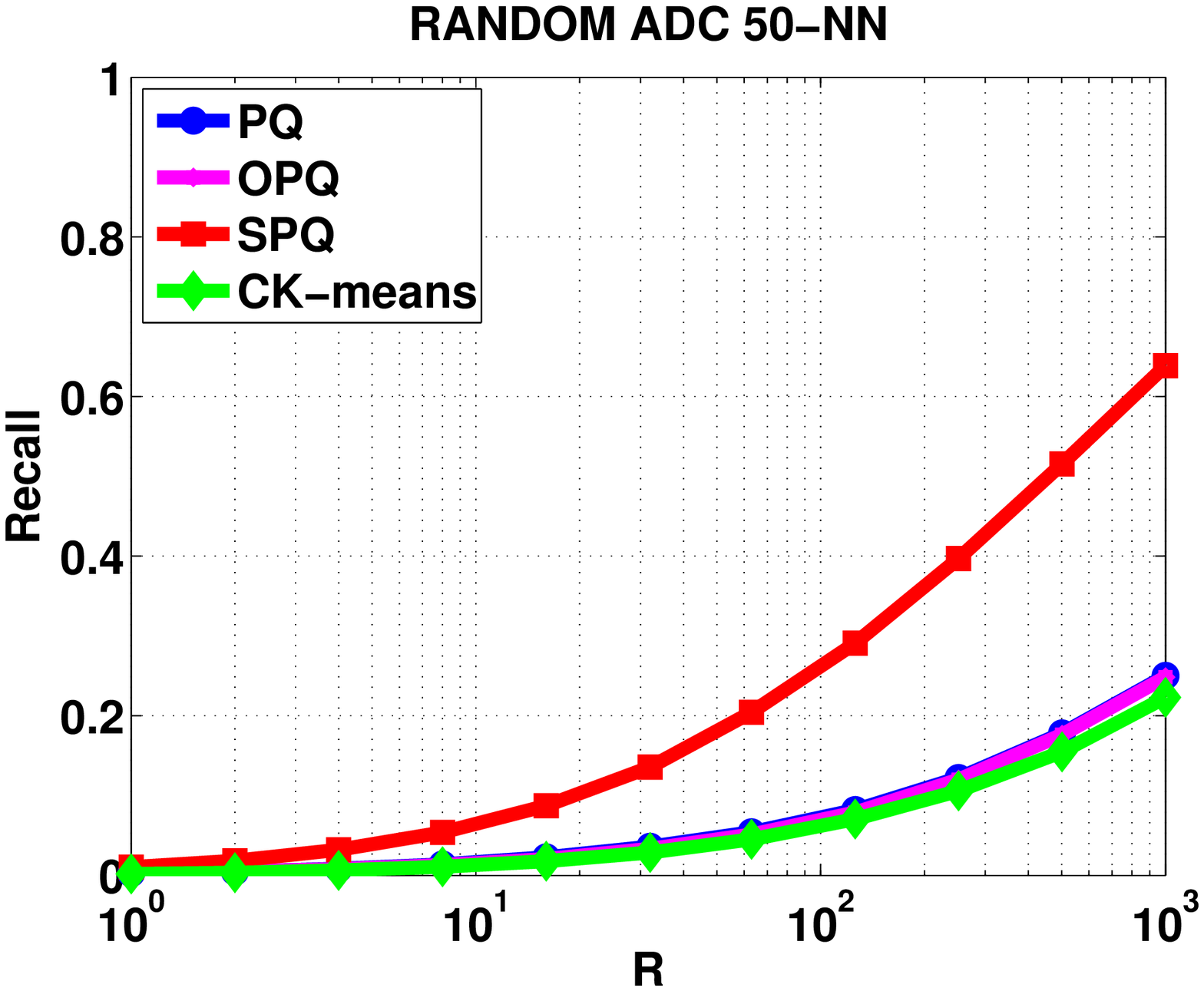}
        \includegraphics[width=\textwidth]{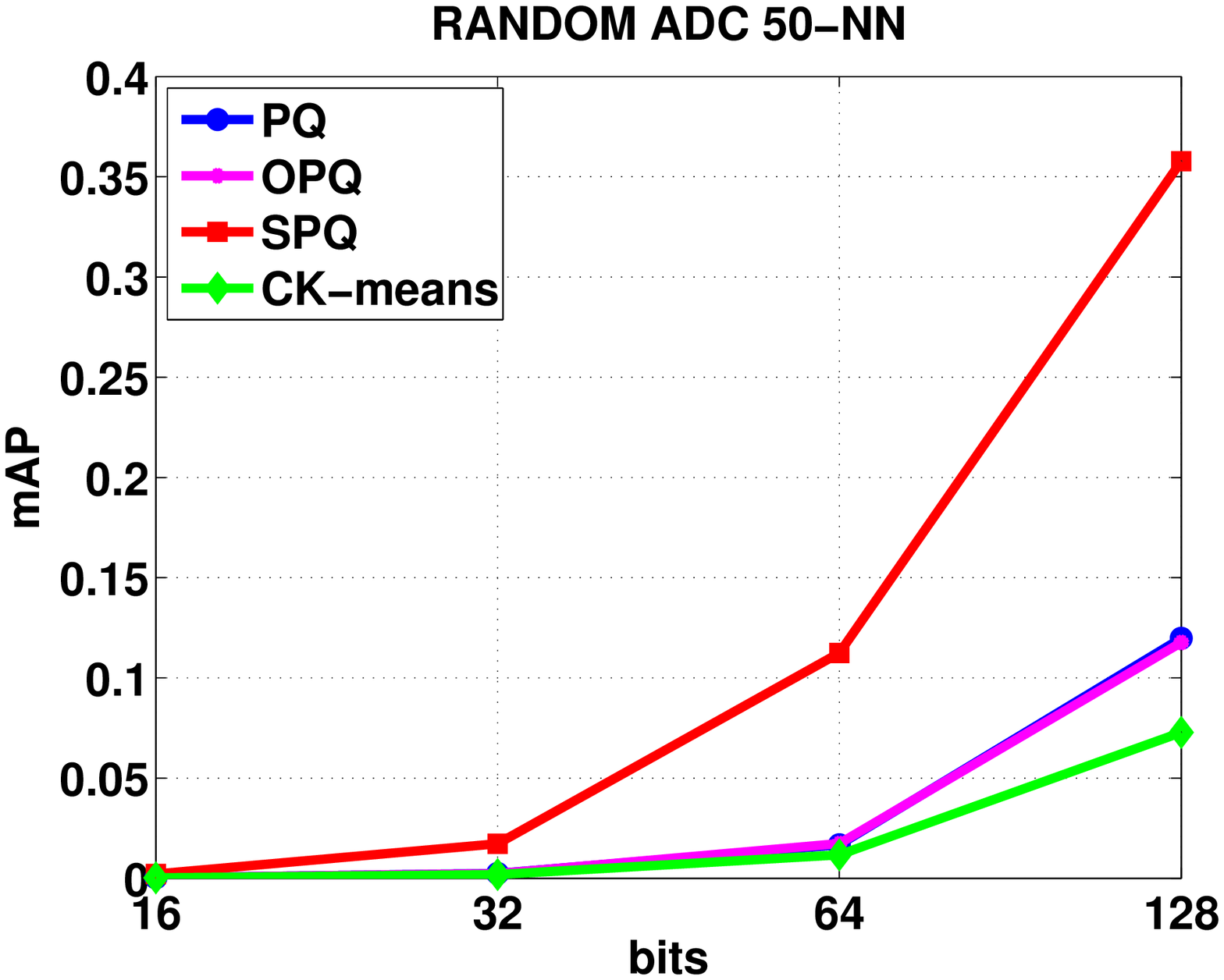}
    \end{minipage}
    \label{fig:random}}%
    \subfloat[]{
    \begin{minipage}{0.24\textwidth}
        \includegraphics[width=\textwidth]{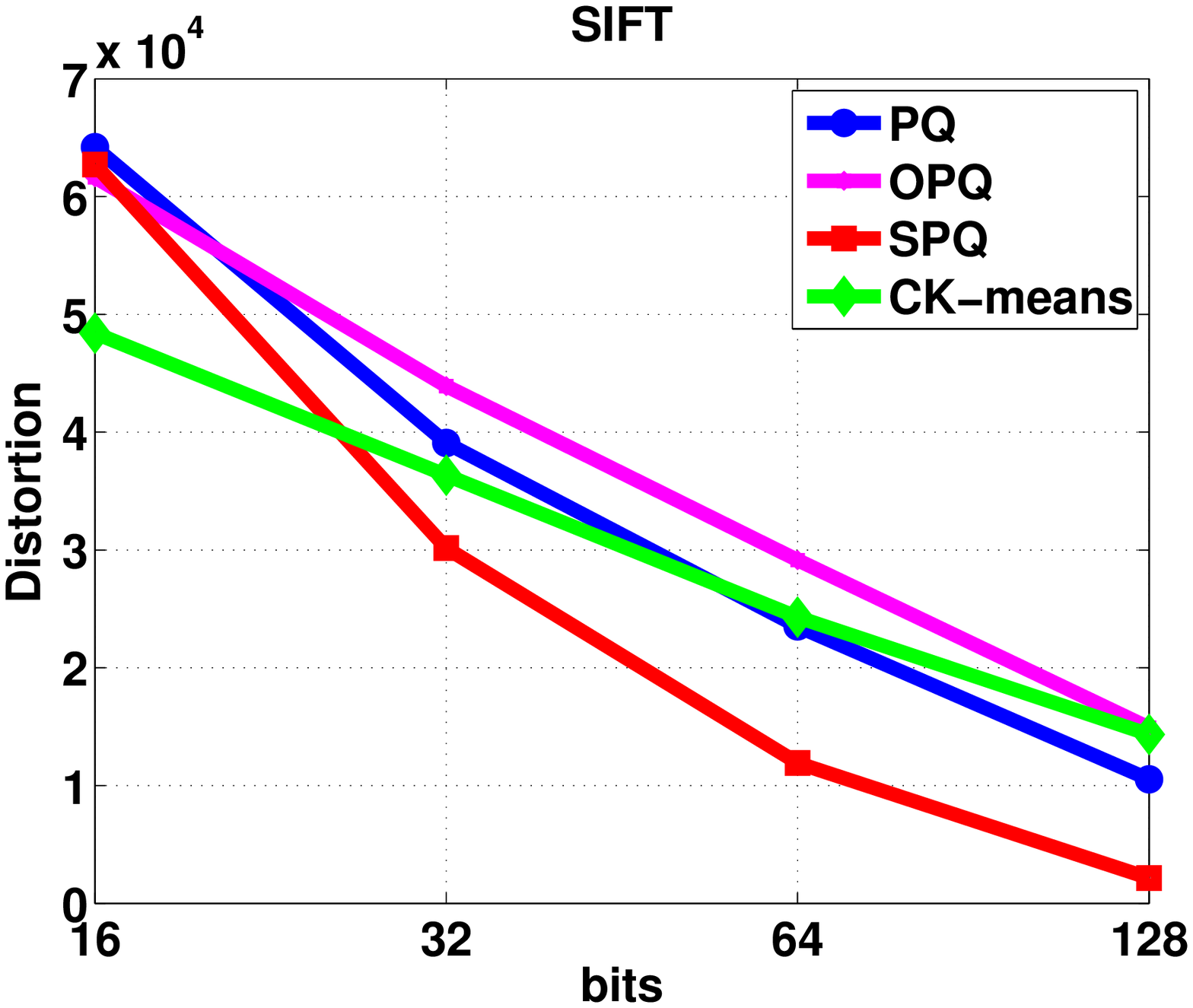}
        \includegraphics[width=\textwidth]{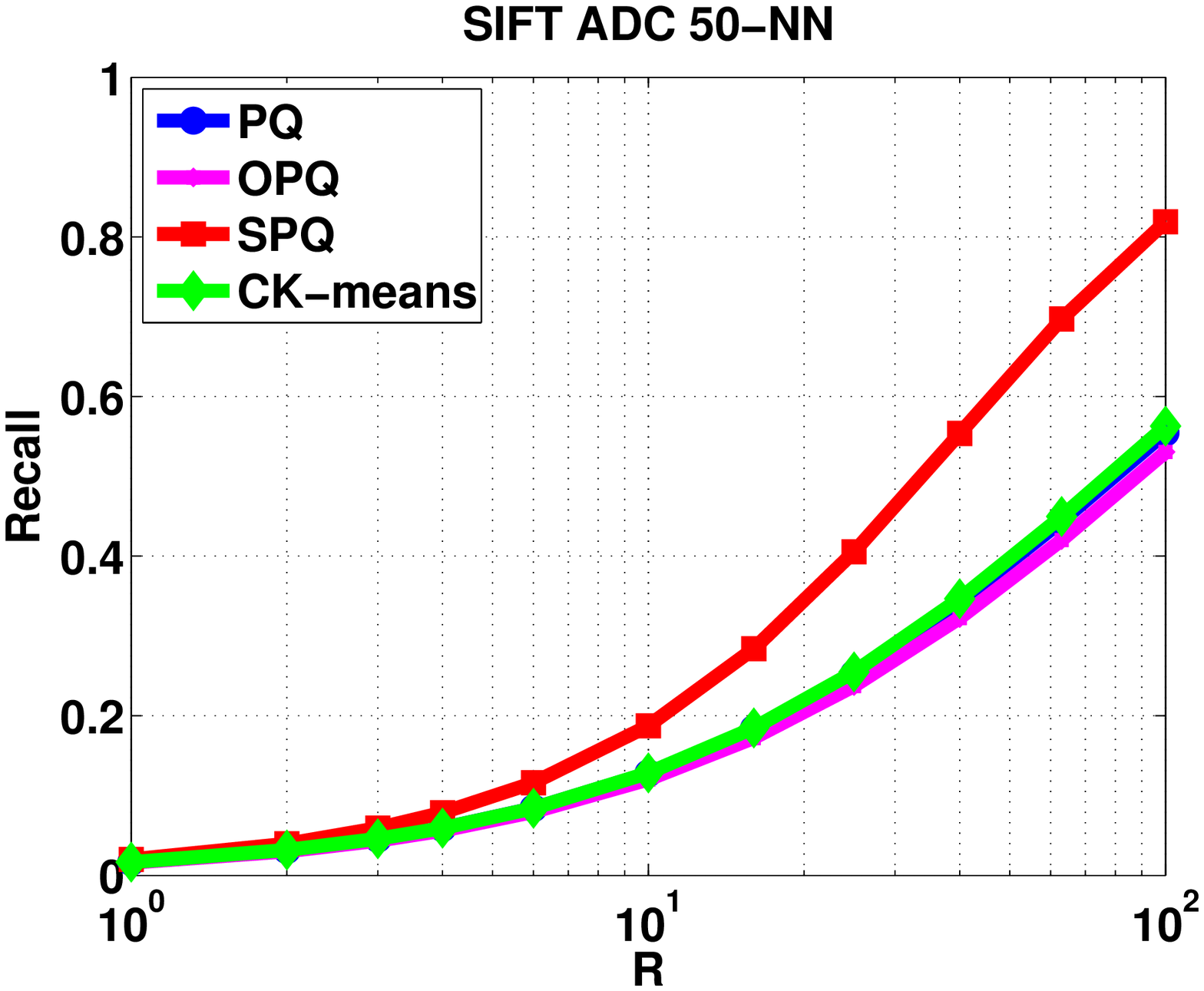}
        \includegraphics[width=\textwidth]{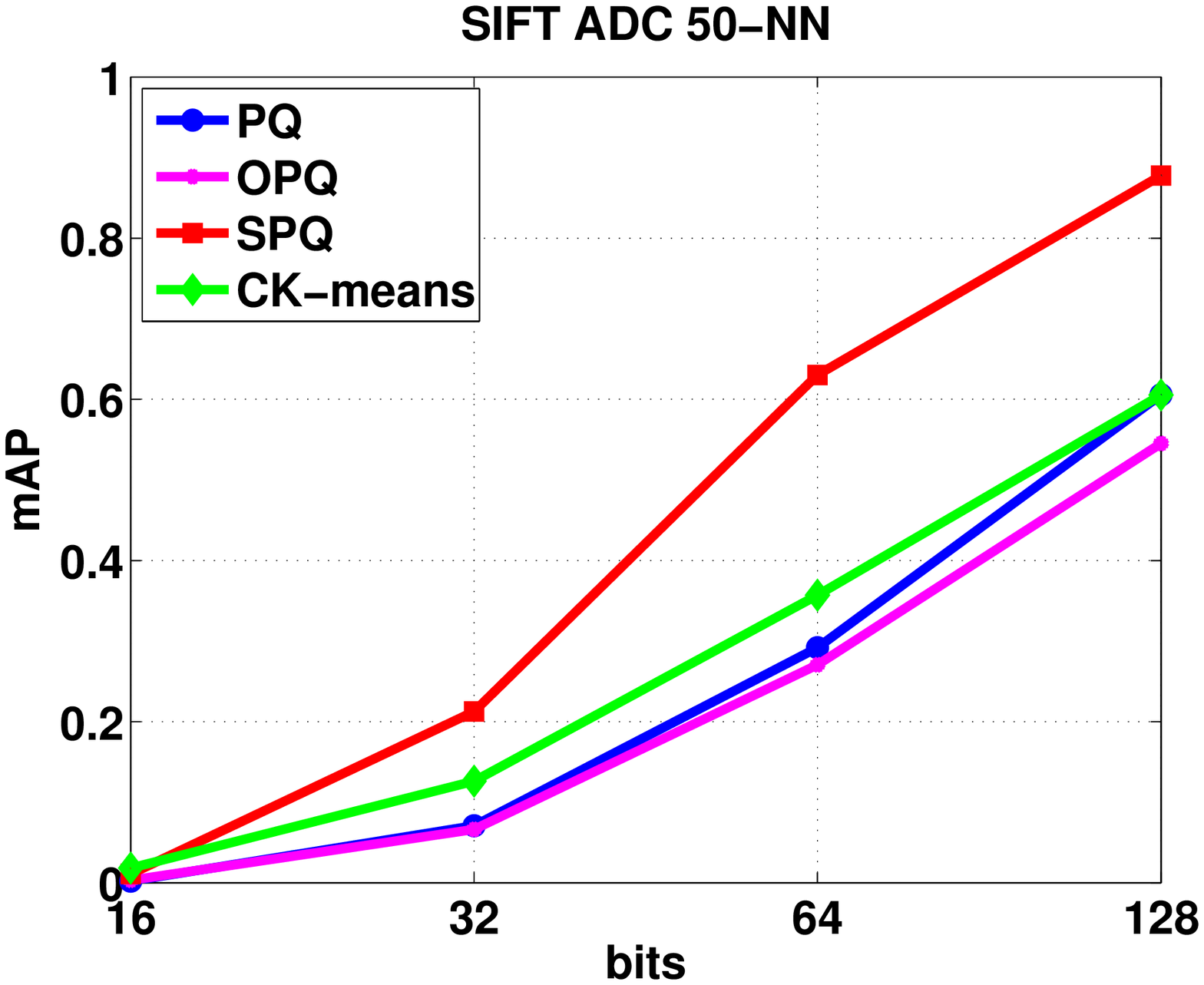}
    \end{minipage}
    \label{fig:sift}}%
    \subfloat[]{
    \begin{minipage}{0.24\textwidth}
        \includegraphics[width=\textwidth]{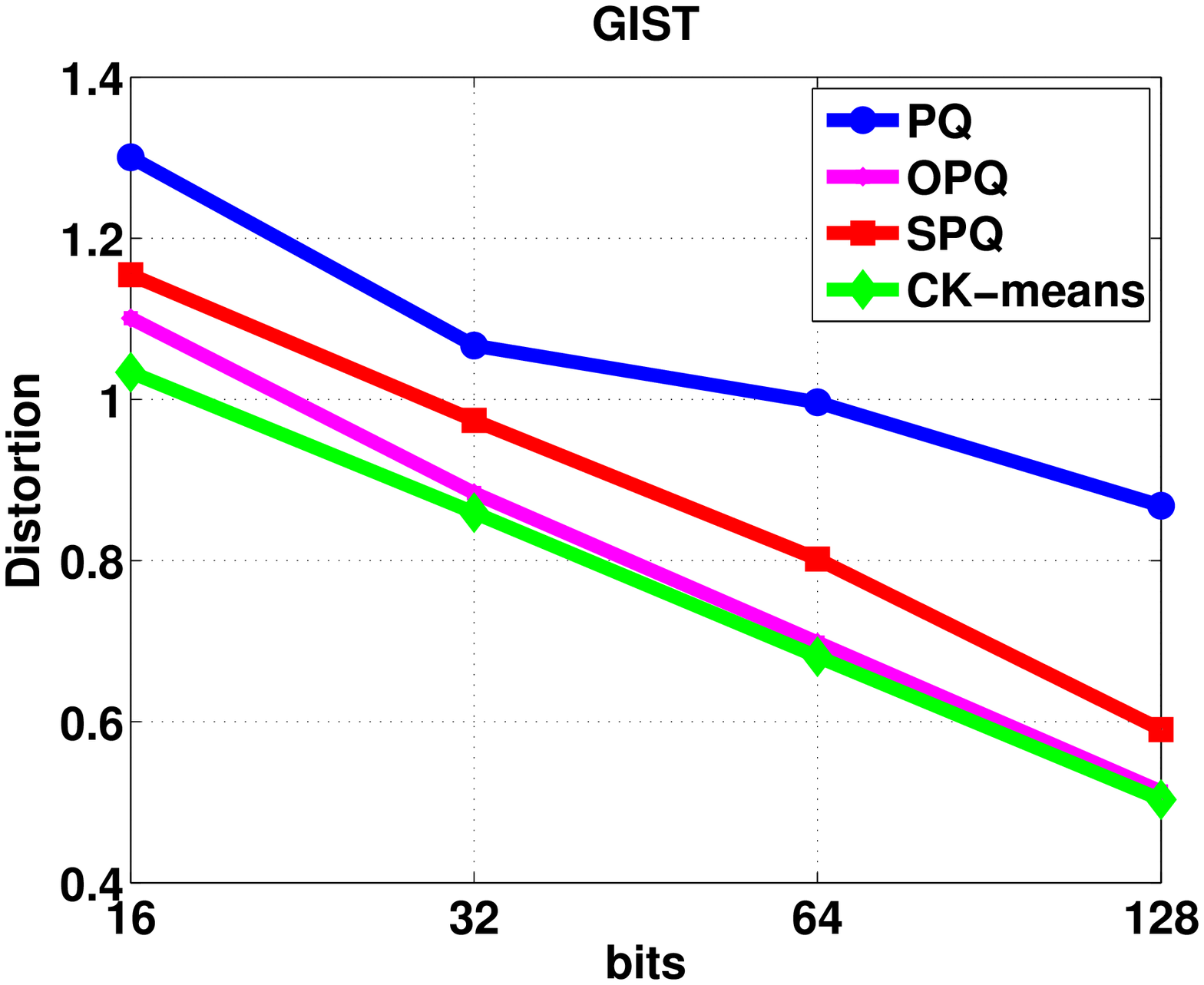}
        \includegraphics[width=\textwidth]{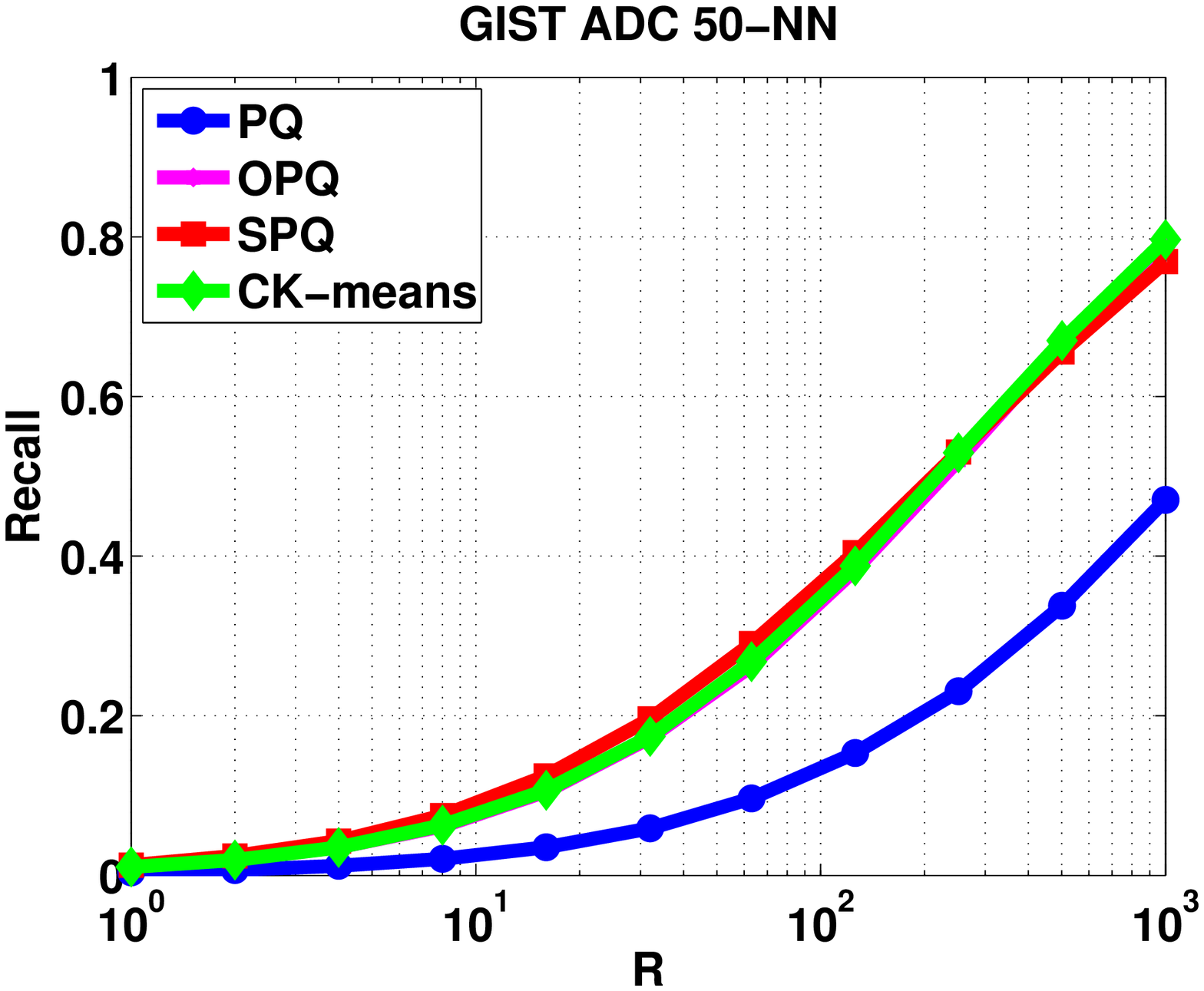}
        \includegraphics[width=\textwidth]{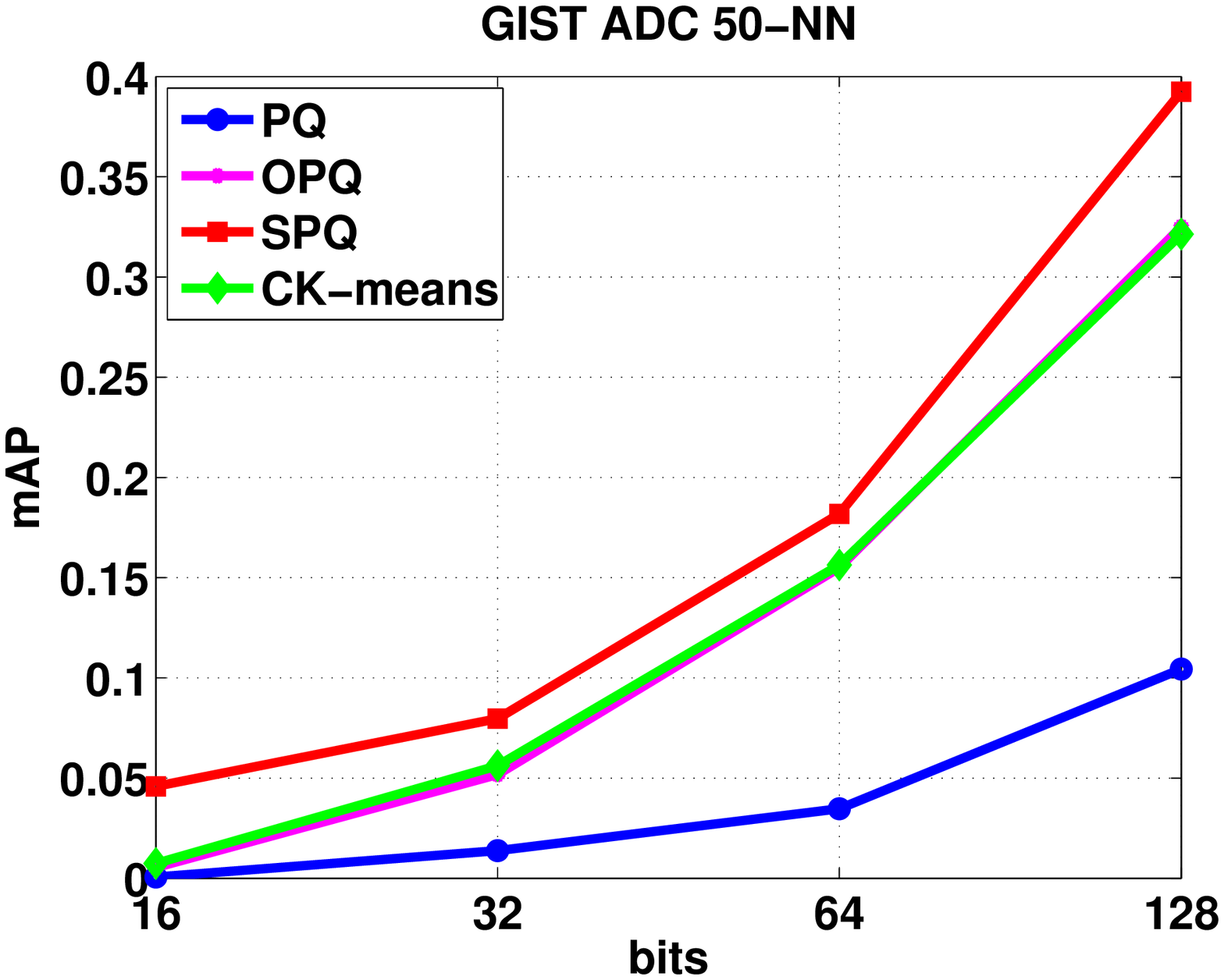}
    \end{minipage}
    \label{fig:gist}}%
    \subfloat[]{
    \begin{minipage}{0.24\textwidth}
        \includegraphics[width=\textwidth]{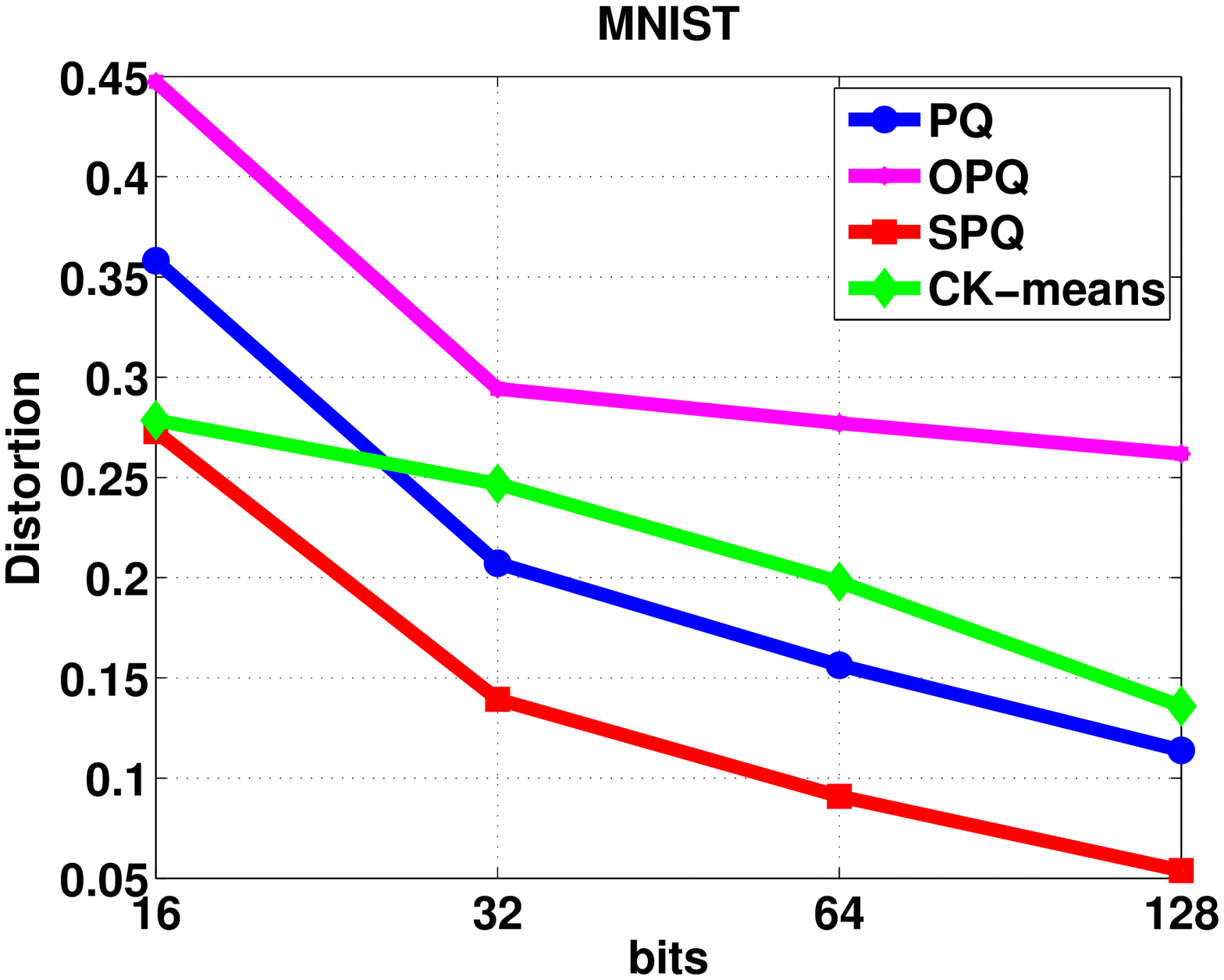}
        \includegraphics[width=\textwidth]{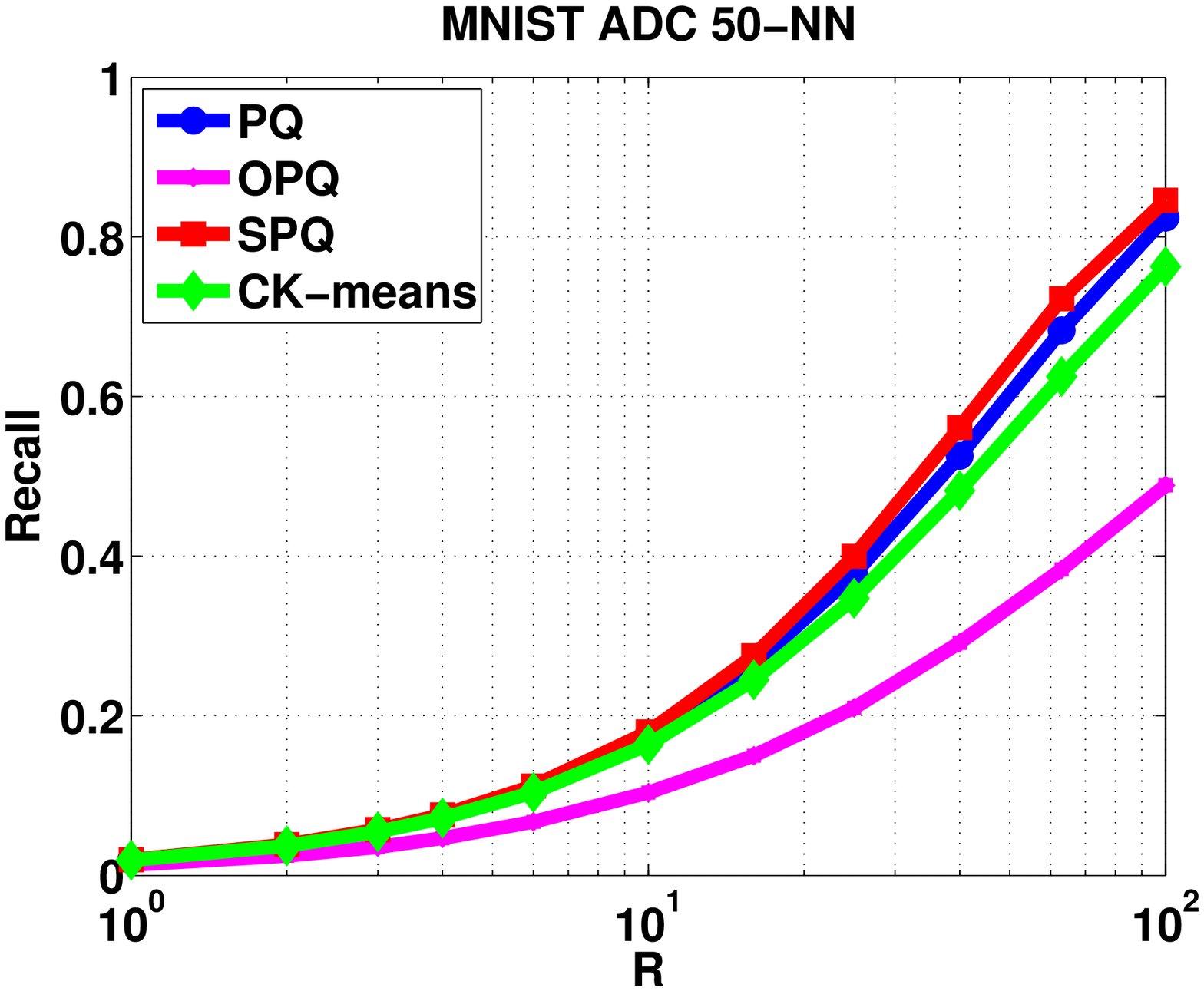}
        \includegraphics[width=\textwidth]{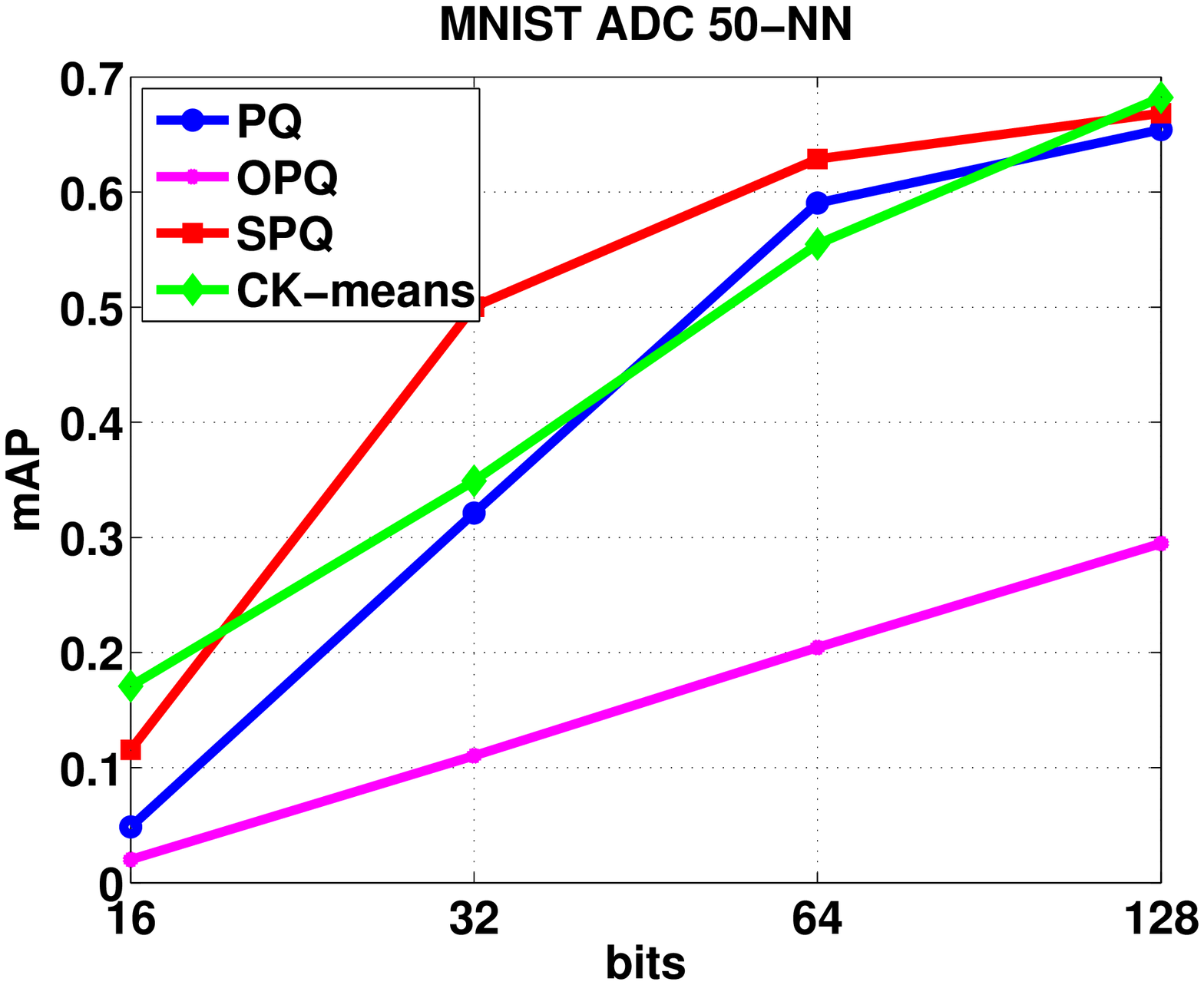}
    \end{minipage}
    \label{fig:labelme}}%
    \caption{
        Performance comparison on different datasets. \protect \subref{fig:random} - \protect \subref{fig:labelme} are the results on dataset Random, SIFT, GIST, and MNIST, respectively.  The first row present the performance in terms of  Distortion vs. Code Length. The second and third row are performance comparison in terms of  Recall vs. R and mAP vs. Code Length ,respectively, when finding the top 50 nearest neighbors. Note that the poor performance of OPQ on MNIST is because we fixed its initilization.}
    \label{fig:distortionbit}
\end{figure*}

\textbf{Codebook Training.}
In what will follow, we study four different kinds of codebook generating methods. They are random sampling, K-means, sparse dictionary learning with $L_1$-norm, and sparse dictionary learning with $L_0$-norm. 
Random sampling method simply generates the codebook via a Gaussian distribution. Both Random method and K-means are learning-free methods. 
The sparse dictionary learning method is based on an $L_1$ constraint or $L_0$ constraint, both of which can be solved by the online algorithm~\cite{Mairal_2009_ODL}. 
We test these codebook methods on SIFT dataset and the results are shown in Fig.~\ref{subfig:codebook} and Fig.~\ref{subfig:codebookdist}. 

From the result, we observe that the performance of random sampling method is much worse than that of other methods, 
since its codebook contains no information of the gallery set. 
Surprisingly, K-means clustering and sparse dictionary learning methods perform very similar, 
which again implies the robustness of our method for different codebook learning algorithms. In the followed experiments, unless explicitly stated, the codebooks are genereated by the sparse method with $L_1$ constraint.

\subsection{Comparisons with other methods}
\label{sec:comp}

\textbf{Comparison with Vector Quantization Methods.}
Note that our proposed SPQ approach is based on the framework of VQ. To facilitate the comprehensive evaluation, we compare our method with three state-of-the-art VQ-based methods, including PQ, OPQ and CK-means.

We firstly examine the quantization distortion for different methods with various code lengths. As shown in Fig.~\ref{fig:distortionbit}, it can be observed that our proposed SPQ method consistently achieves very low squared distortion on all the datasets compared to the other methods. Then, we evaluate the performance in terms of recall vs. R when searching for  different numbers of NNs, and also measure the recall with respect to the total number of returned candidates, i.e., recall vs. R in the result. Moreover, we also measure the recall@100, which denotes the proportion of true NNs in the top 100 returned NN results with various code lengths. Based on whether or not quantizing the queries, there are two different kinds of distance computation methods: ADC and SDC. Fig.~\ref{fig:adcsdc} shows the experimental results. From the results, it is clear to see that our approach generally outperforms the other competing quantization methods.

As presented in \cite{JDS11}, PQ slightly improves the search accuracy by combining an inverted file structure and encoding the residual. We also utilize the inverted file structure~(IVFSPQ) and compares with it~(IVFPQ) in Fig.~\ref{fig:adcsdc}. We can see that IVFPQ indeed perform slightly better than PQ in both the SIFT and GIST dataset, while our method still outperform both of them.
For the efficiency, our proposed SPQ approach is expected to be slightly more computational expensive than PQ. However, as shown in Table~\ref{tb:speed2}, the empirical time costs for PQ and SPQ when using an inverted file structure~(IVFPQ vs. IVFSPQ) are fairly comparable.


    
\begin{figure*}[htbp]
    \centering
    \subfloat[]{
    \begin{minipage}{0.24\textwidth}
        \includegraphics[width=\textwidth]{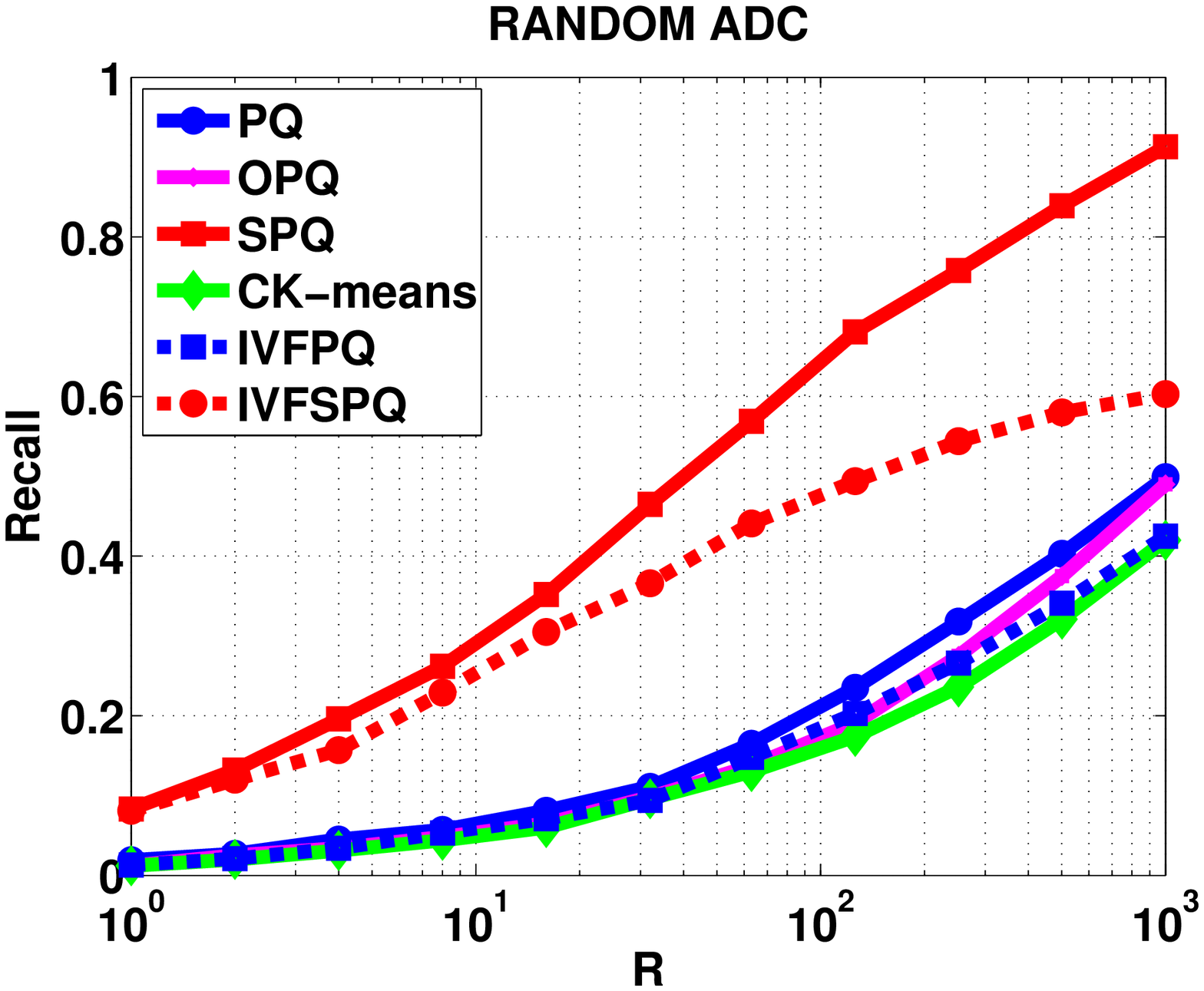}
        \includegraphics[width=\textwidth]{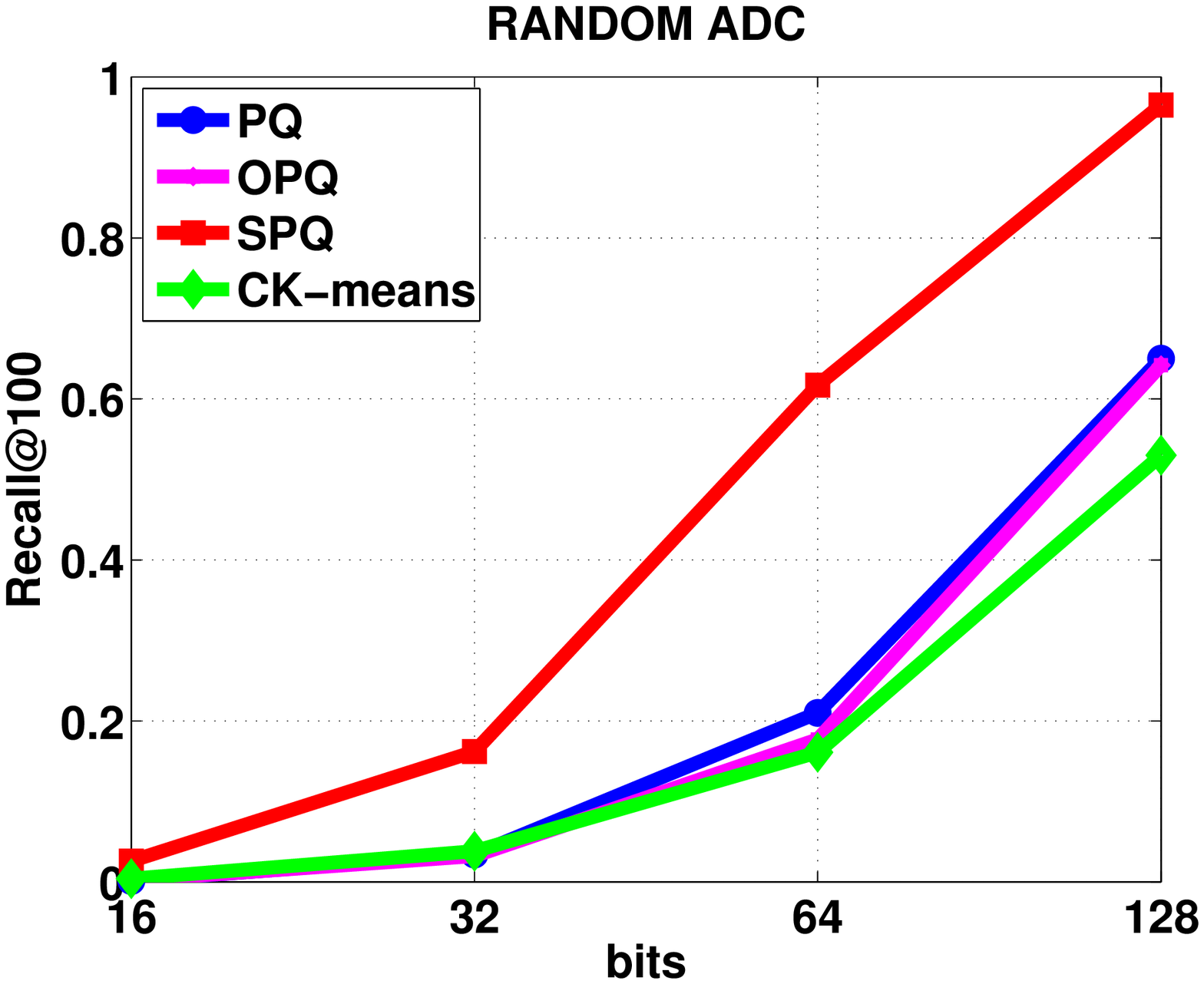}
        \includegraphics[width=\textwidth]{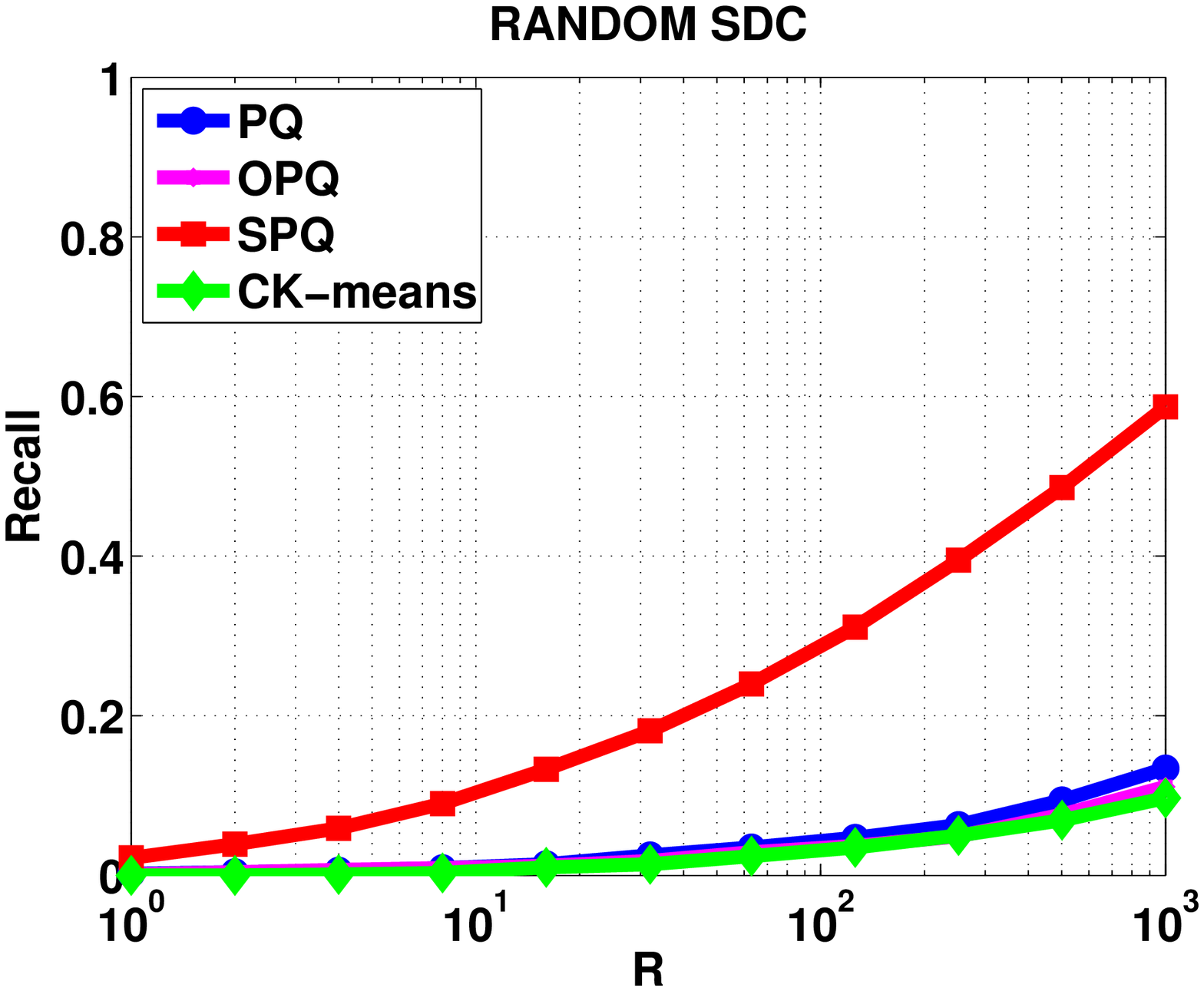}
        \includegraphics[width=\textwidth]{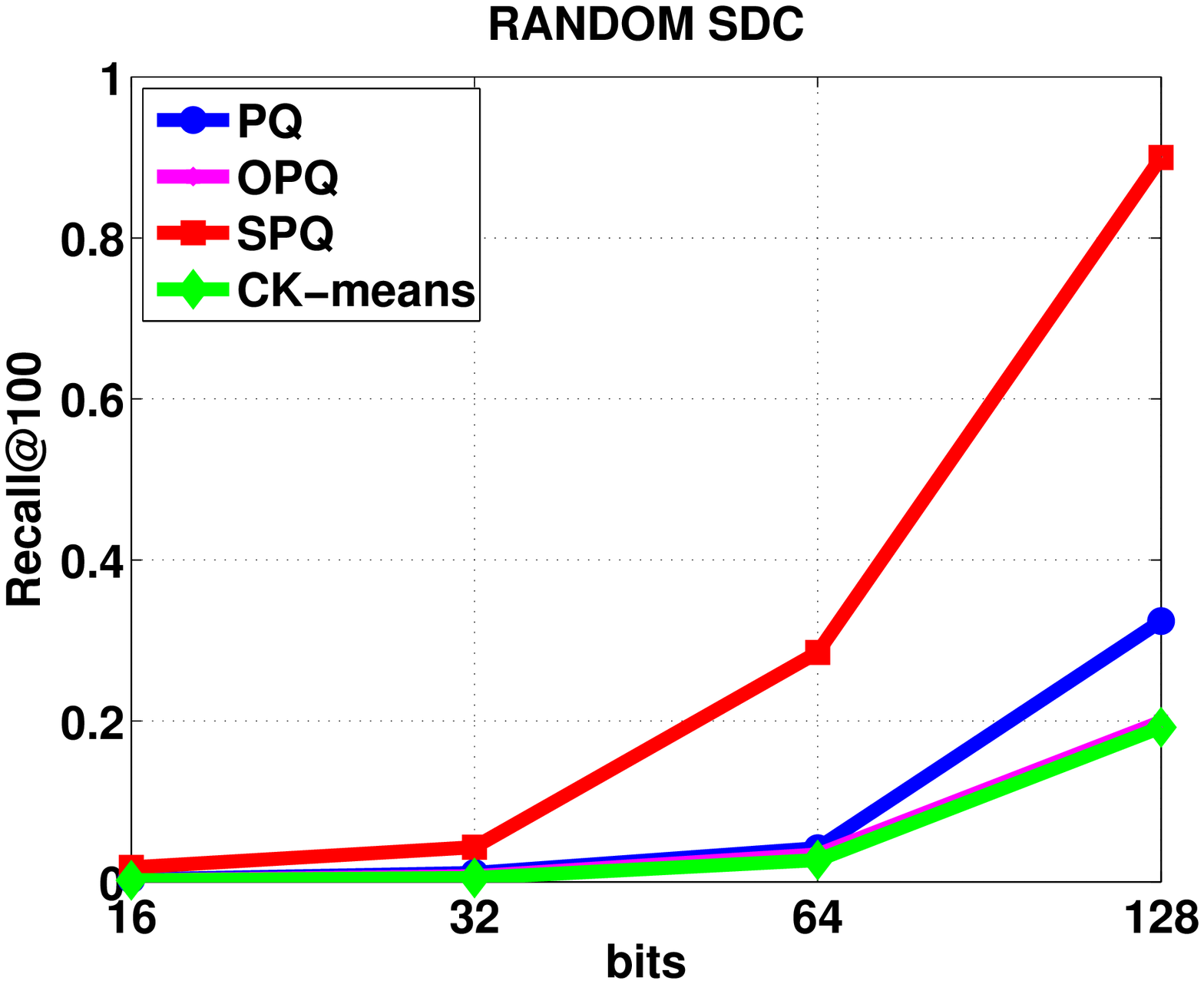}
    \end{minipage}
    \label{subfig:random}}%
    \subfloat[]{
    \begin{minipage}{0.24\textwidth}
        \includegraphics[width=\textwidth]{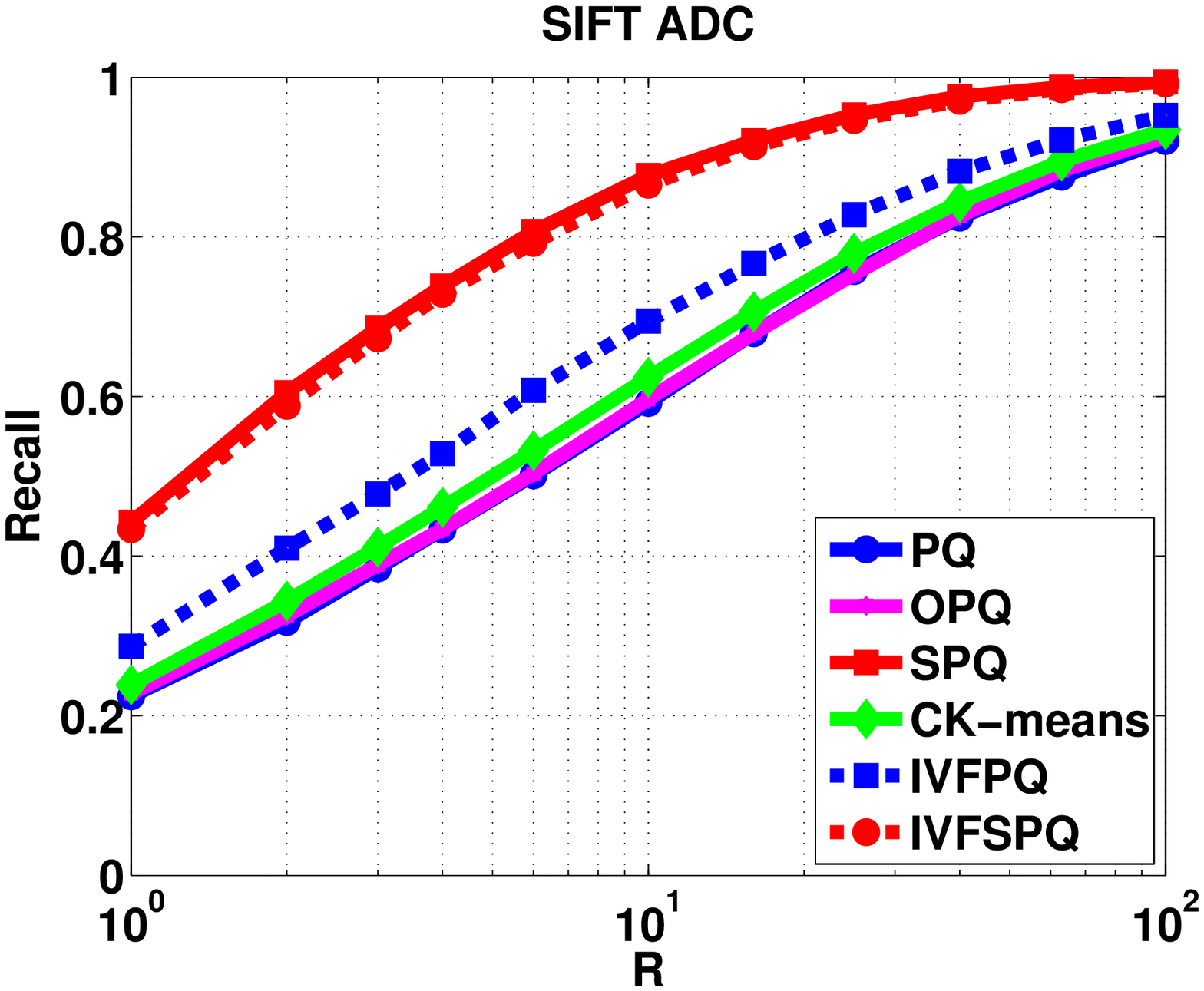}
        \includegraphics[width=\textwidth]{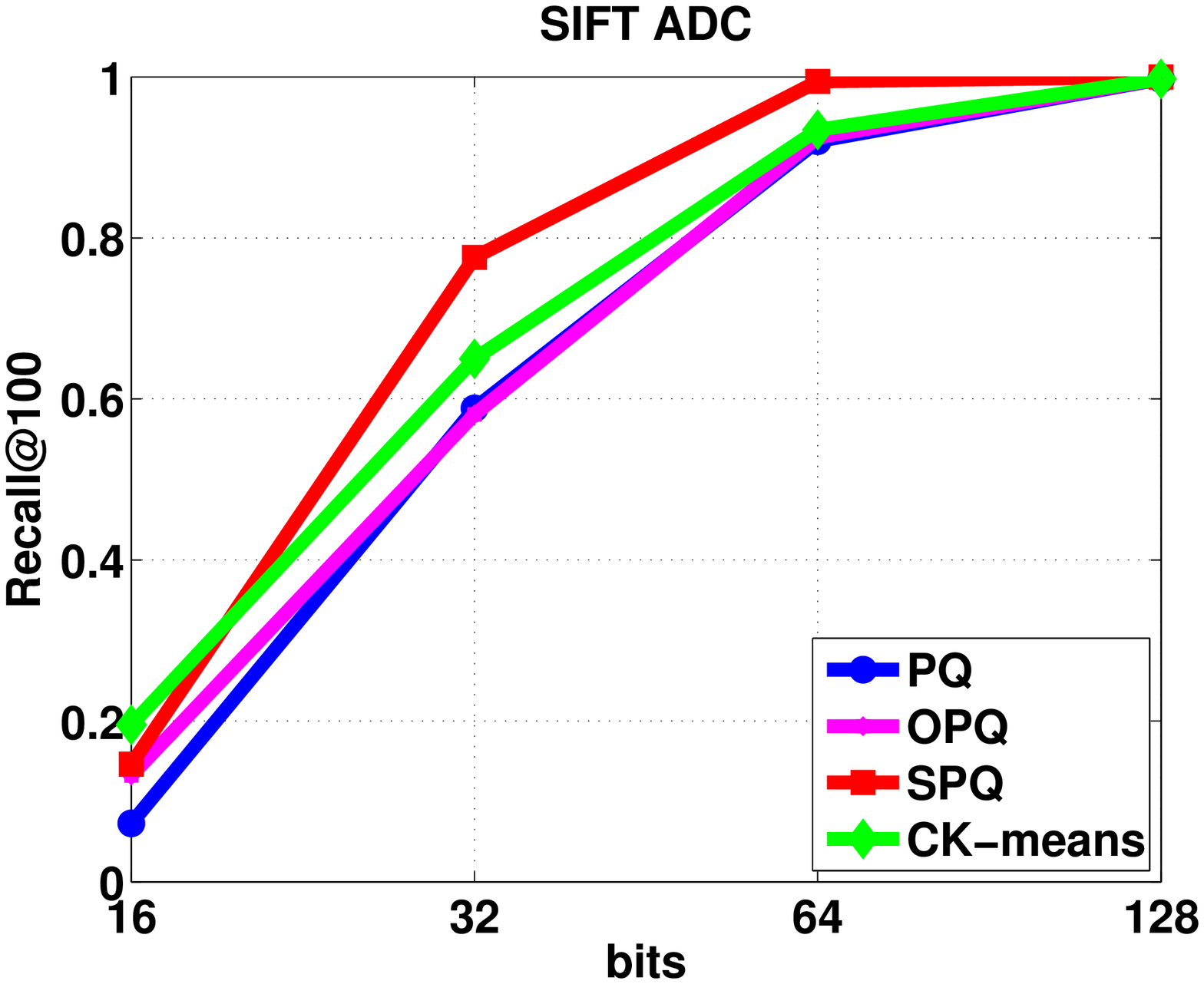}
        \includegraphics[width=\textwidth]{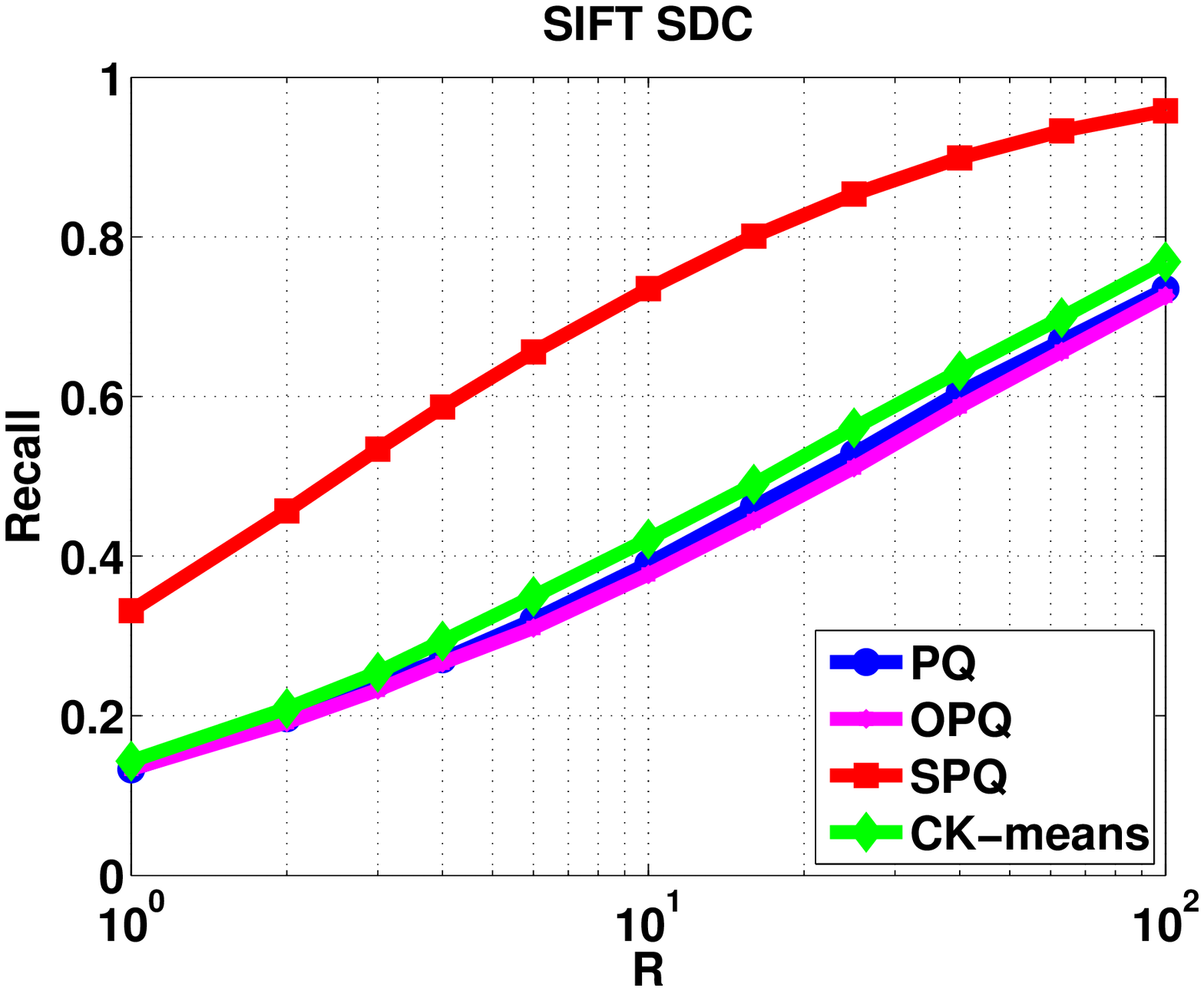}
        \includegraphics[width=\textwidth]{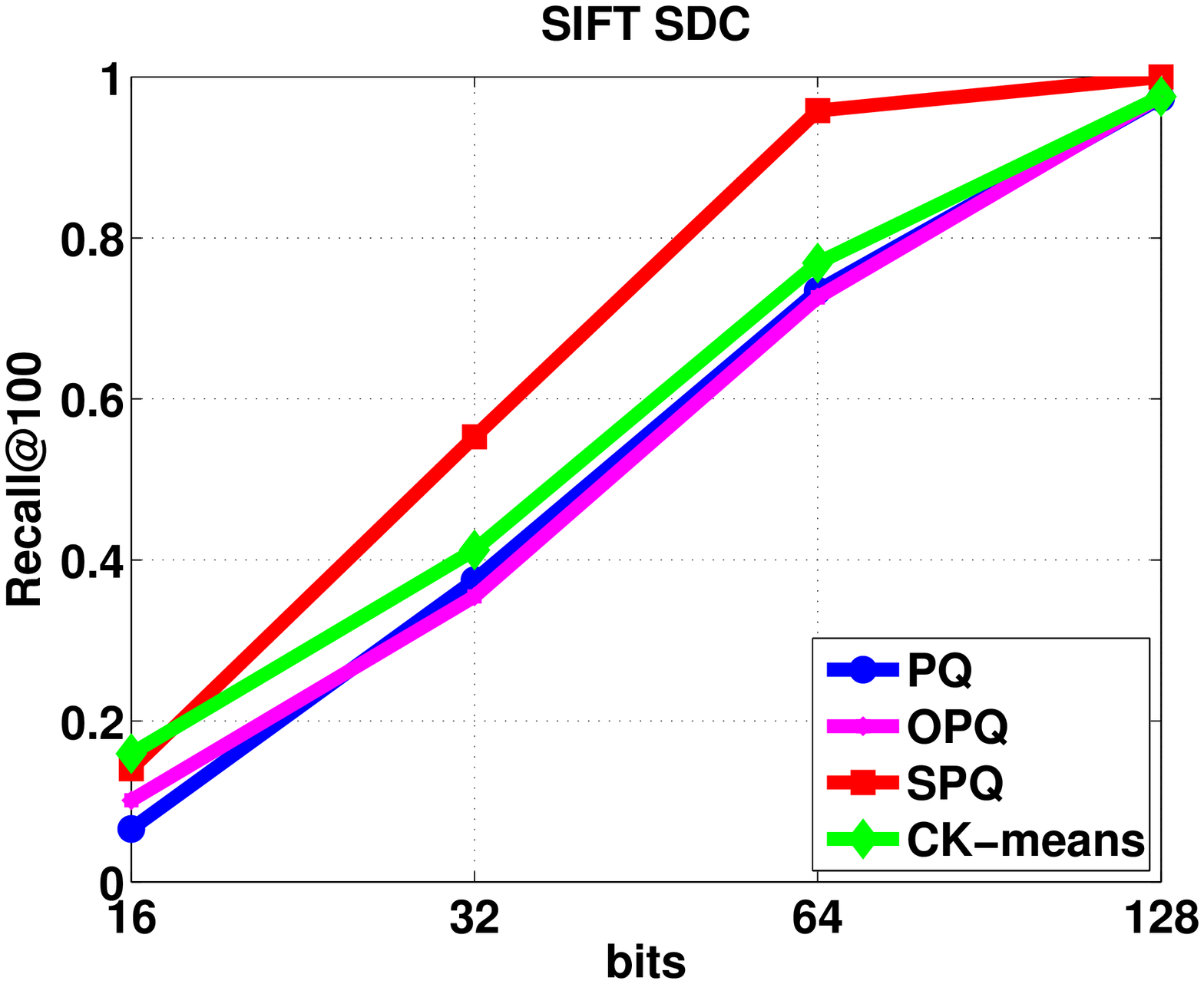}
    \end{minipage}
    \label{subfig:sift}}%
    \subfloat[]{
    \begin{minipage}{0.24\textwidth}
        \includegraphics[width=\textwidth]{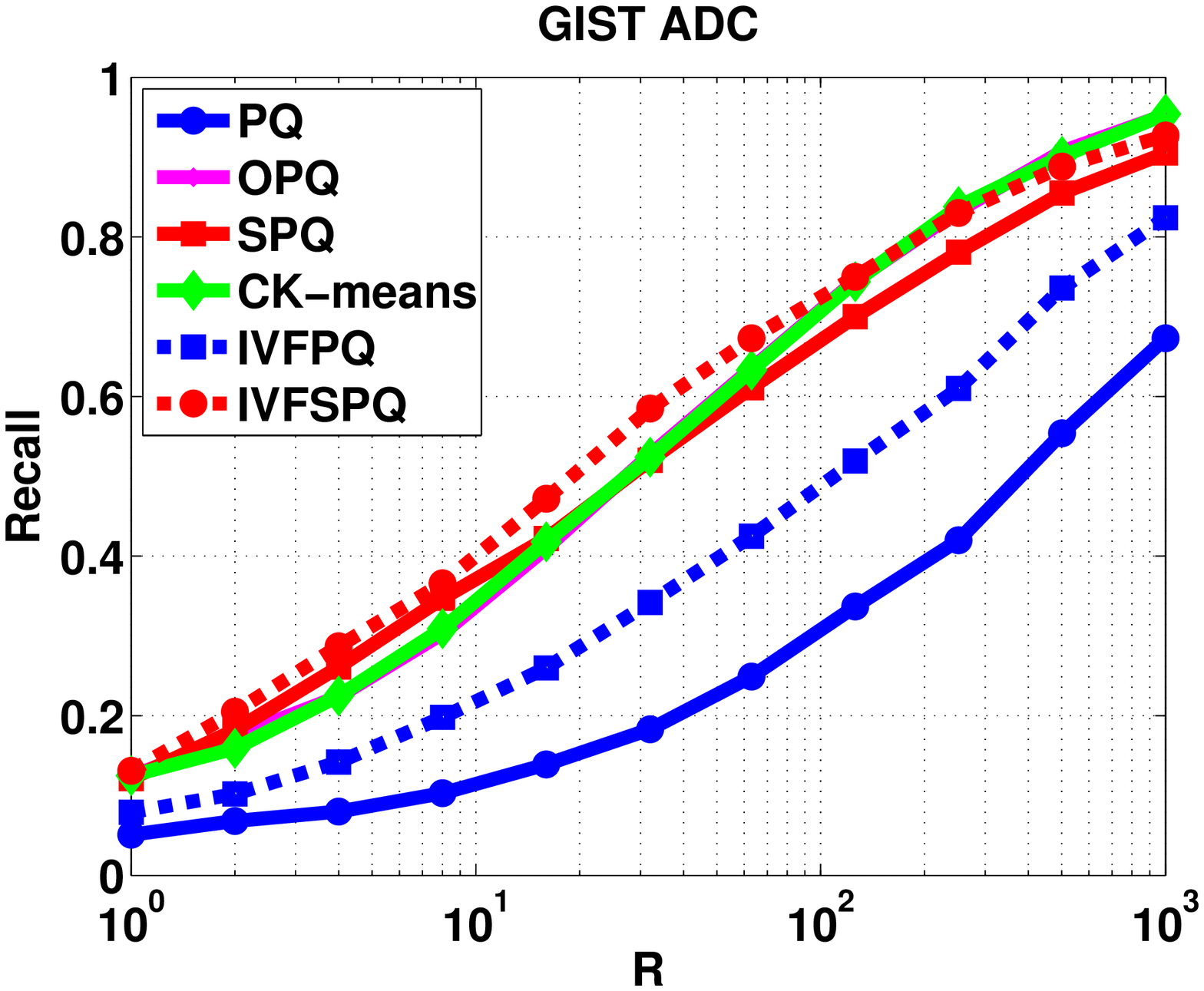}
        \includegraphics[width=\textwidth]{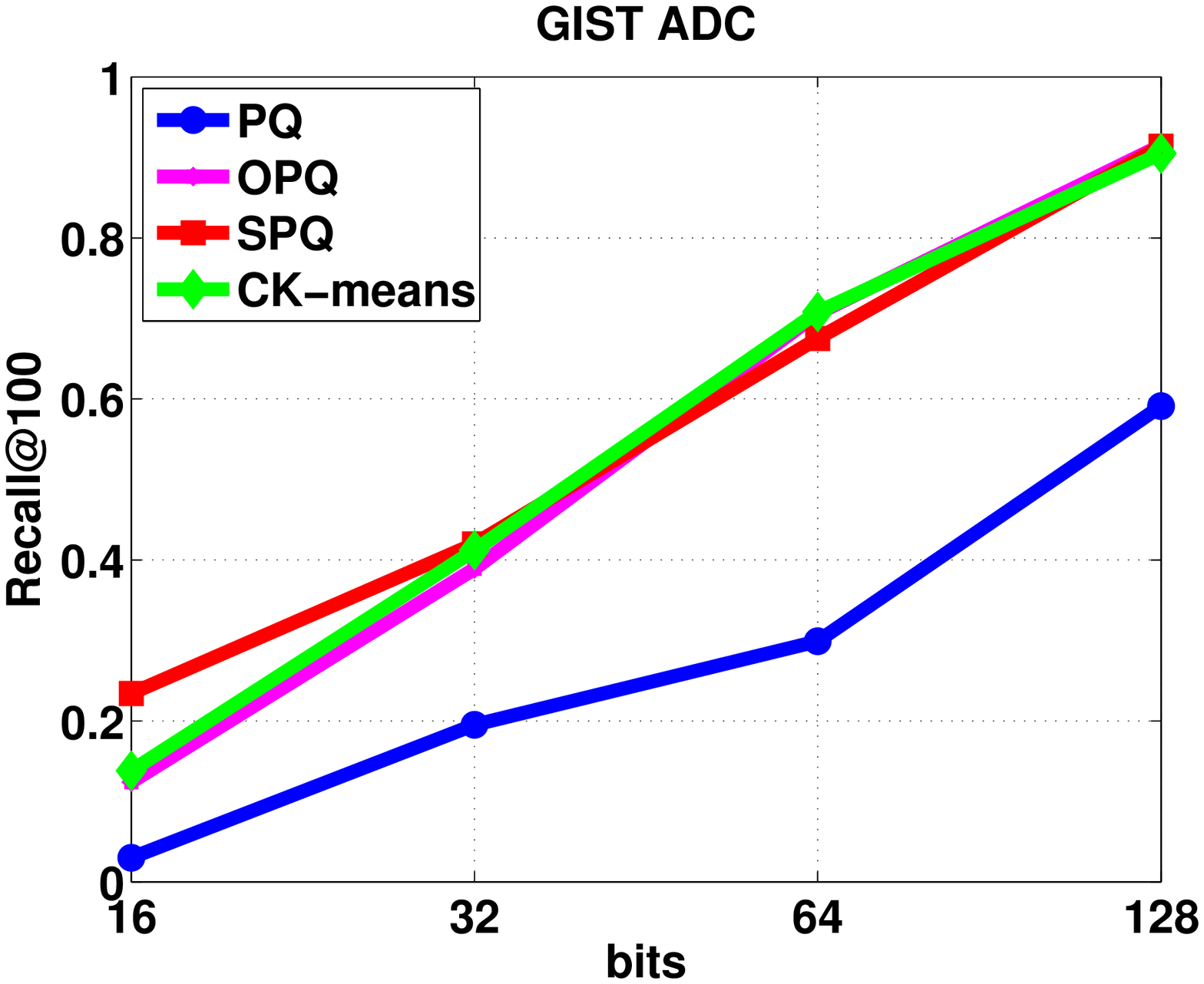}
        \includegraphics[width=\textwidth]{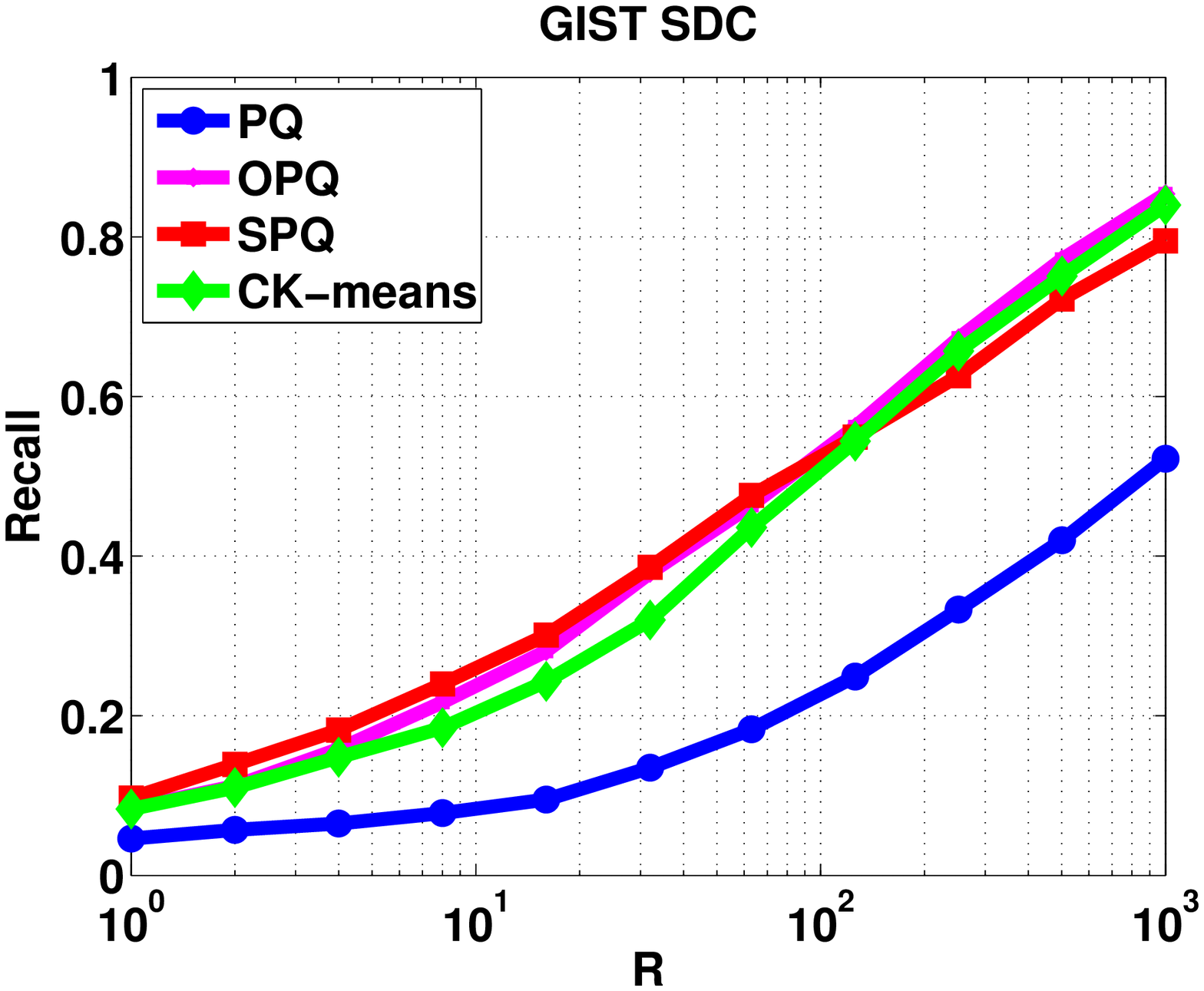}
        \includegraphics[width=\textwidth]{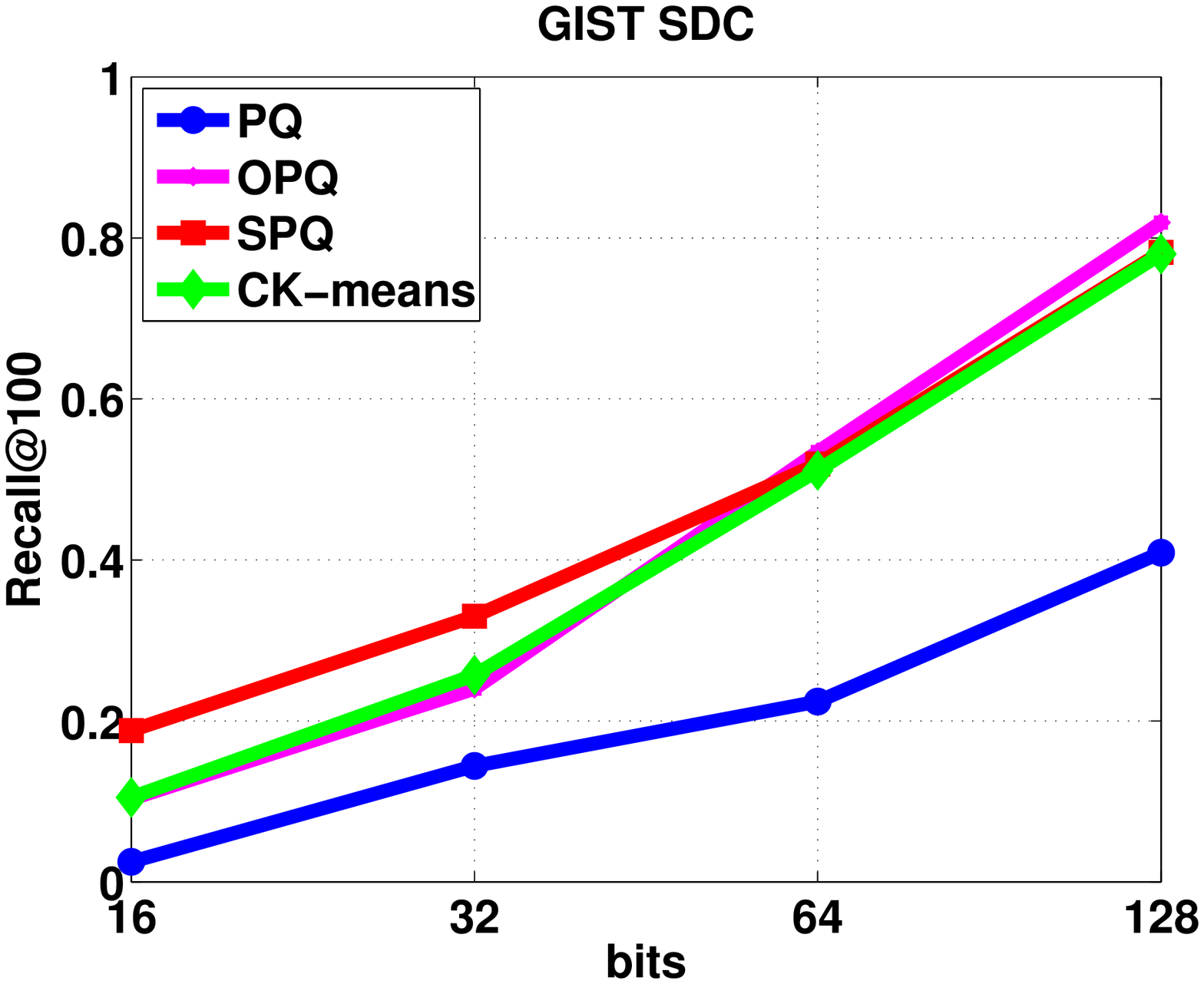}
    \end{minipage}
    \label{subfig:gist}}%
    \subfloat[]{ \begin{minipage}{0.24\textwidth}
        \includegraphics[width=\textwidth]{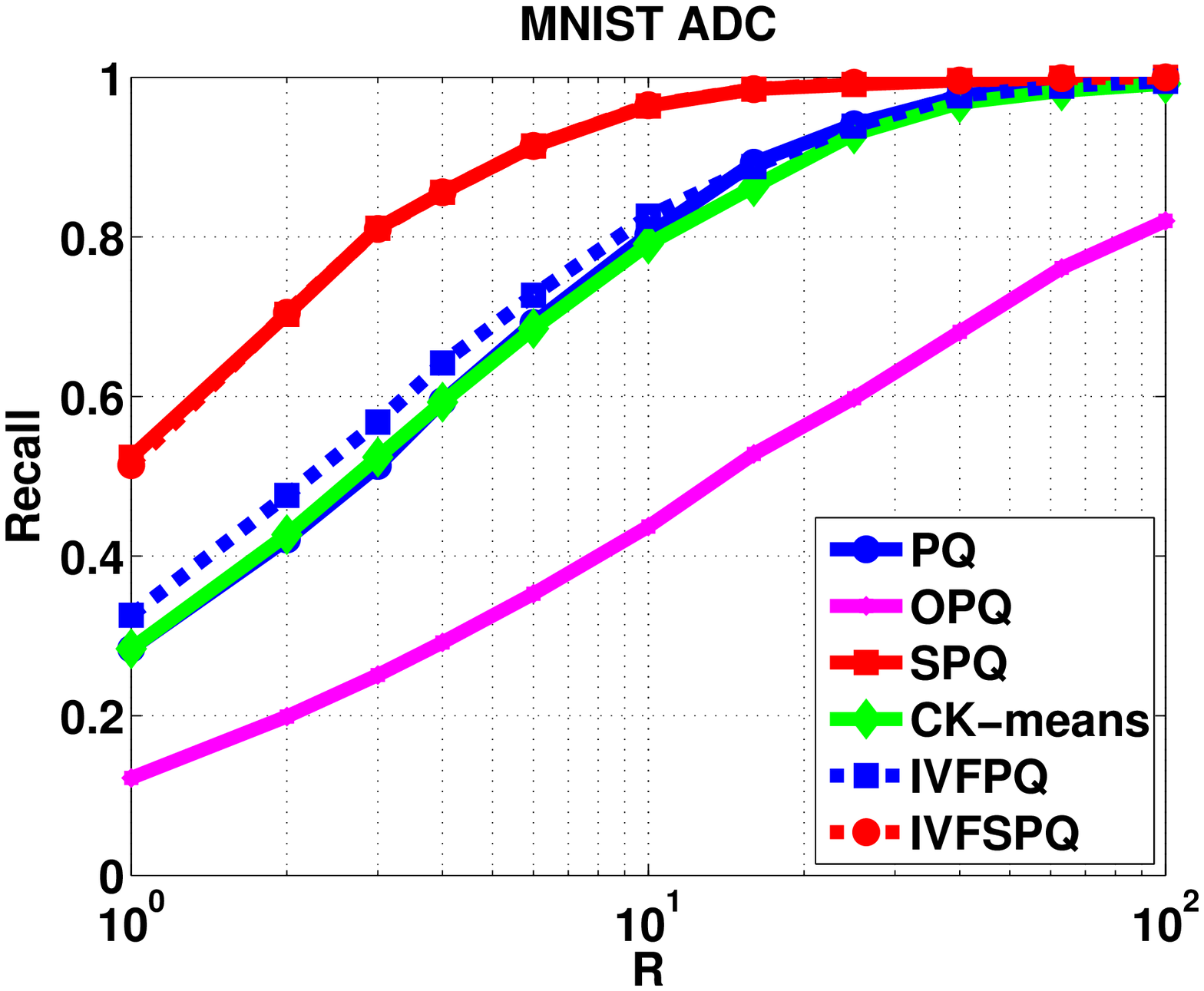}
        \includegraphics[width=\textwidth]{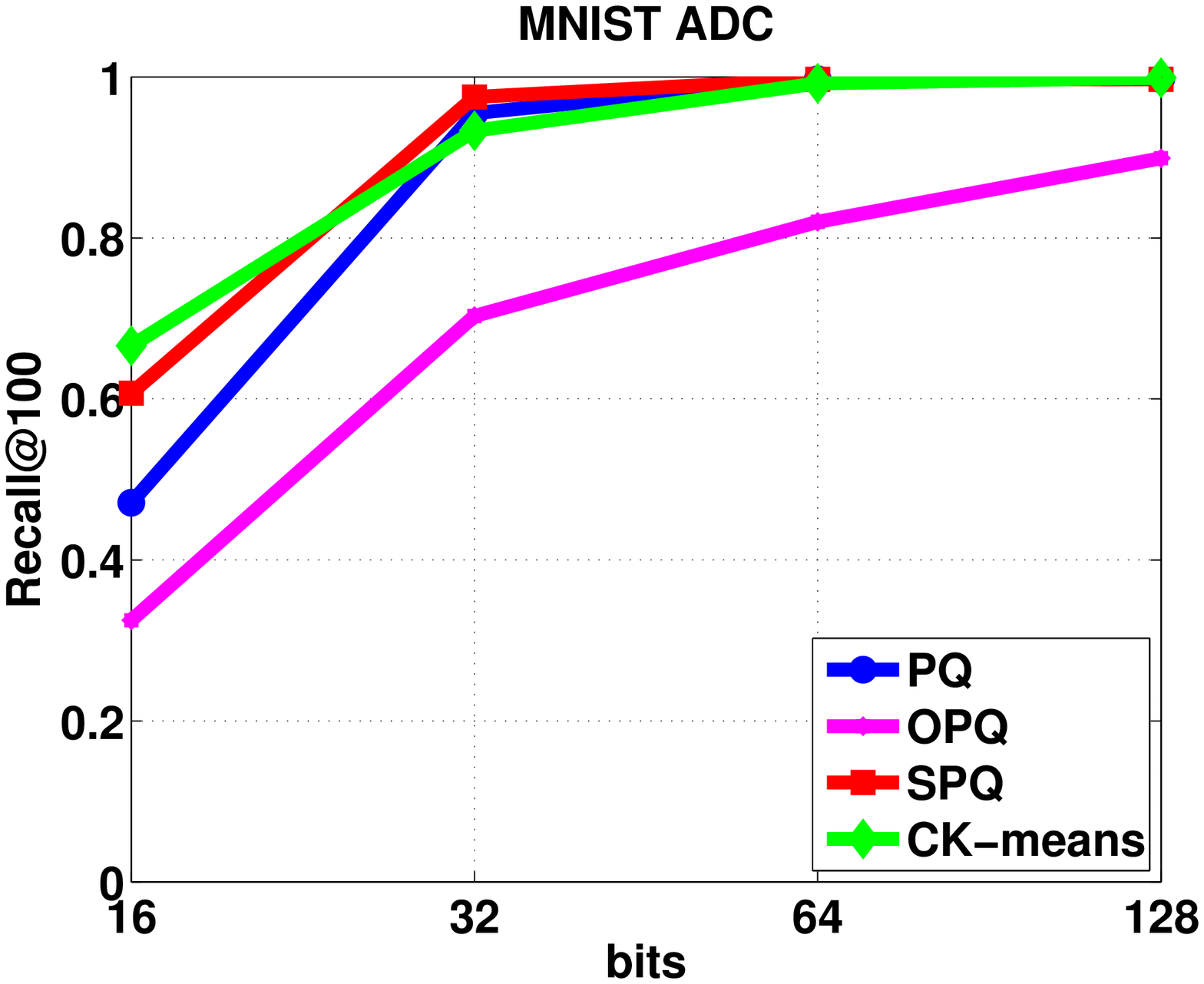}
        \includegraphics[width=\textwidth]{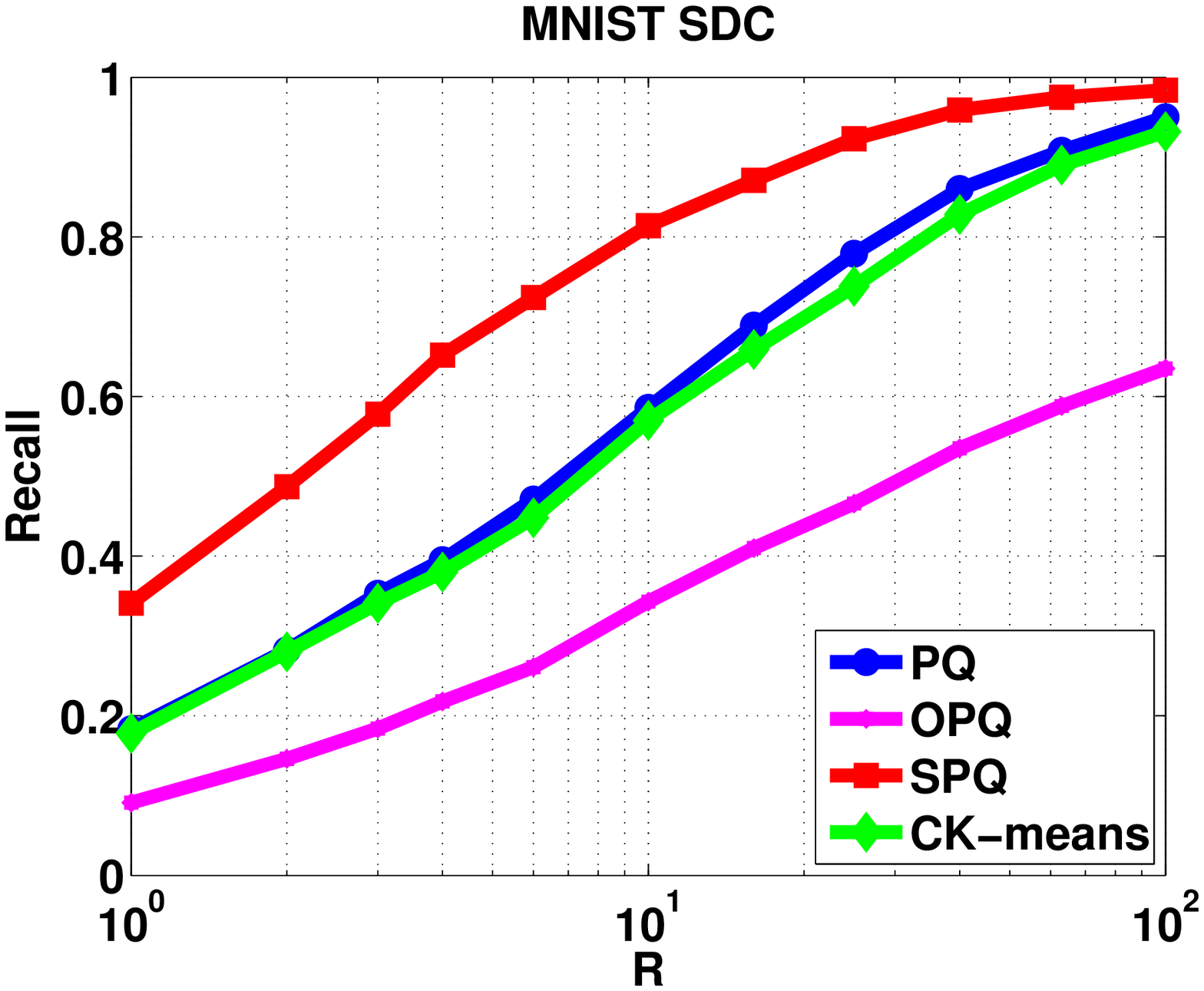}
        \includegraphics[width=\textwidth]{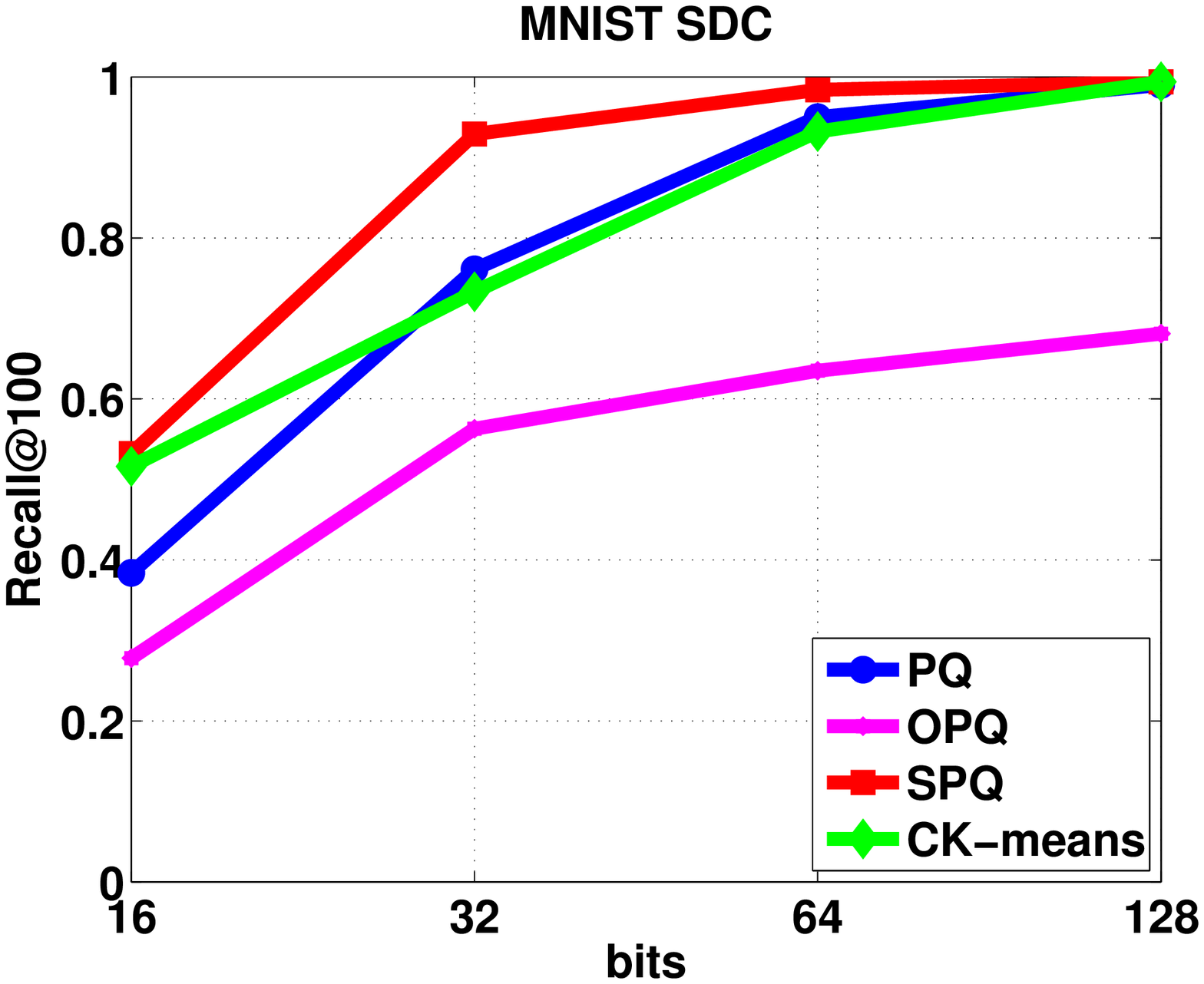}
    \end{minipage}
    \label{subfig:labelme}}%
    \caption{ Performance comparison of ADC and SDC on different datasets. \protect \subref{fig:random} - \protect \subref{fig:labelme} are the results on dataset Random, SIFT, GIST, and MNIST, respectively.
     Performance comparison are in terms of  Recall vs. R and mAP vs. Code Length when finding the nearest neighbor. The top two rows are under ADC and the bottom two are under SDC. In the first row, we also compare between PQ and SPQ that employ the inverted file structure, i.e. IVFPQ and IVFSPQ. It can be observed that our method performs better, if not competitive, than other methods.}
    \label{fig:adcsdc}
\end{figure*}

\textbf{Comparison with Hashing-based Methods.}

We compare our SPQ approach with several state-of-the-art hashing-based ANN search methods, including minimal loss hashing~(MLH)~\cite{norouzi2011minimal}, iterative quantization hashing~(ITQ)~\cite{ITQ}, order preserving  hashing~(OPH)~\cite{conf/mm/WangWYL13}, locality sensitive hashing~(LSH)~\cite{LSH}, kernelized supervised hashing~(KSH)~\cite{liu2012supervised}, isotropic hashing~(IsoHash)~\cite{kong2012isotropic}, and spectral hashing~(SH)~\cite{weiss2009spectral}. 
To make a fair comparison, we follow the evaluation protocol in~\cite{conf/mm/WangWYL13}, 
where mAP is employed as performance metric   with the ground truth being 50 nearest neighbors.
Table~\ref{tb:hashing} shows the performance evaluation on three datasets, 
including LabelMe, SIFT and GIST. It can be clearly seen that our proposed SPQ approach significantly outperforms these hashing-based methods at a large margin.
    To make it clear, we compare searching time with the spectral hashing algorithm~(SH~\cite{weiss2009spectral}), and the results are summarized in Table~\ref{tb:speed2}. For PQ~\cite{JDS11}, we have re-implemented the Hamming distance computation in C in order to ensure that all the approaches in our comparisons are optimized appropriately . It can be seen that our proposed method outperforms SH in terms of both efficiency and accuracy.


\begin{table*}[htbp]
    \centering
    \caption{Comparison with Hashing Methods (mAP).  Our approach obtains a significant gain over all of these methods on all the datasets. }
    \label{tb:hashing}
    \begin{tabular}{c|c|cccccccc}
        \Xhline{2.5\arrayrulewidth}
        \multirow{2}{*}{Dataset} & \multirow{2}{*}{\vspace{0.13in}Code} & \multicolumn{8}{c}{Approaches}\\
        \cline{3-10}
        & Length & SPQ & MLH~\cite{norouzi2011minimal} & ITQ~\cite{ITQ} & OPH~\cite{conf/mm/WangWYL13} & LSH~\cite{LSH} & KSH~\cite{liu2012supervised} & IsoHash~\cite{kong2012isotropic} & SH~\cite{weiss2009spectral}\\
        \Xhline{2.5\arrayrulewidth}
        \multirow{3}{*}{LabelMe}
                                 & 32 & \textbf{47.97} & 19.91 & 20.36 & 21.11 & 8.87 & 16.72 & 18.51 & 9.28\\
                                 & 64 & \textbf{62.14} & 32.48 & 32.09 & 33.94 & 17.57 & 24.57 & 28.35 & 11.18\\
                                 & 128 & \textbf{77.52} & 45.22 & 44.66 & 44.36 & 32.52 & 31.45 & 42.34 & 13.73\\
        \hline
        \multirow{3}{*}{SIFT}
                              & 32  &\textbf{ 32.08}& 3.07 & 2.69 & 5.07  &  1.49 & 1.26 & 2.31 & 4.23\\
                              & 64  &\textbf{ 69.47}&  8.11 & 8.16 & 13.58 & 5.68 & 2.94 & 7.24 & 9.81\\
                              & 128 &\textbf{ 86.39} &  18.01 & 17.87 & 26.00 & 14.05 & 5.04 & 16.53 & 15.56\\
        \hline
        \multirow{3}{*}{GIST}
                              & 32 &\textbf{ 4.07 } & 1.74 & 1.68 & 2.00  & 0.56 & 1.15 & 1.39 & 0.68\\
                              & 64 &\textbf{ 6.40 } & 3.51 & 3.27 & 4.12  & 1.50  & 2.21 & 3.25 & 1.08\\
                              & 128&\textbf{ 11.14} & 5.96 & 5.14 & 6.97 & 3.12 &3.83 & 5.21 & 1.45\\
        \Xhline{2.5\arrayrulewidth}
    \end{tabular}
\end{table*}

\textbf{Comparison with Searching Tree-based Method.}

It is interesting to compare our proposed SPQ approach with FLANN, which is known as the most popular open-source toolbox for ANN search. 
We select the SIFT dataset as the testbed, and evaluate the precision with given searching time. 
As in PQ~\cite{JDS11}, we take advantage of an inverted file structure to speed up the SPQ method with the cost of slight performance drop. 
FLANN includes a re-ranking scheme that computes the exact distances for the candidate nearest neighbors.
For the sake of comparison with FLANN, we also add a re-ranking stage to our SPQ method. In practice, while obtaining a precision of $94.7\%$, SPQ costs $6.0$ seconds or $4.2$ seconds employing SSE in the search stage. FLANN, however, takes $6.4$ seconds to obtains a precision of $84.2\%$.
Fig.~\ref{fig:flann} shows the experimental results. The precision of our method is obtained by re-ranking 50 returned candidates and the extra time cost is less than 0.1 second. 
It is not difficult to see that our method is faster than FLANN if the precision of re-ranking results is required to be higher than $0.7$. This is critical since we always pursue higher precision given a fixed period of time.
More importantly, our method consumes much less memory cost than the searching tree-based FLANN method: the indexing structure occupies less than 100MB, while FLANN requires more than 250 MB of RAM. We should notice that our result could be further improved with a better inverted method, such as multi-index~\cite{multi_index}.


\begin{figure}[tbp]
    \centering
    \includegraphics[width=0.4\textwidth]{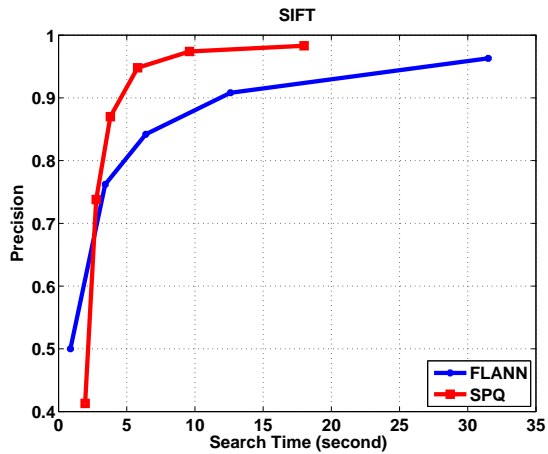}%
    \caption{Comparison with FLANN~\cite{flann_pami_2014}. The precision of our method is obtained by re-ranking 50 returned candidates and the extra time cost is less than 0.1 second. 
        It can be clearly seen that our method is faster than FLANN if the precision of re-ranking results is above $0.7$.
}
    \label{fig:flann}
\end{figure}

\begin{table*}[htbp]
    \centering
    \caption{
            Computational cost and accuracy performance on the SIFT dataset using 64 bits. For better illustration, we also include comparisons of MATLAB implementation with the optimized C++/Mex code in the search stage. 
For the efficiency, our proposed SPQ approach is expected to be slightly more computational expensive than PQ. However, the empirical time costs for PQ and SPQ when using an inverted file structure~(IVFPQ vs. IVFSPQ) are fairly comparable.
For other ANN methods, our method outperforms both SH and FLANN on the search time and precision. The source codes of all the methods are publicly available. 
}
    \label{tb:speed2}
    \begin{tabular}{|c|c|c|cc|}
        \Xhline{2.5\arrayrulewidth}
        \multicolumn{2}{|c}{ Approaches} \vline & Search time~(ms/per) & Recall@1~(\%) & Recall@100~(\%)\\
        \Xhline{2.5\arrayrulewidth}
        \multirow{5}{*}{MATLAB/mex}
             & PQ~\cite{JDS11} & 8.8 & 23.0 & 92.3\\
             & SPQ & 21.9 & 51.9 & 99.8\\
             & IVFPQ~\cite{JDS11} & 1.3 & 26.6 & 92.1\\
             & IVFSPQ & 1.4 & 51.2 & 95.7\\
             & SH~\cite{weiss2009spectral} & 2.2 & 9.5 & 53.0\\
        \hline
        \multirow{2}{*}{C/C++}
             & IVFSPQ & 0.4 & 43.5 & 94.7\\
            & FLANN~\cite{flann_pami_2014} & 0.6 & - & 84.2\\
        \Xhline{2.5\arrayrulewidth}
    \end{tabular}
\end{table*}

\begin{table}[t]
    \centering
    \caption{The retrieval result in the Oxford dataset}
    \label{tb:retrieval}
    \begin{scriptsize}
    \begin{tabular}{c|c|c|c}
        \Xhline{2.5\arrayrulewidth}
        accuracy& SPQ & FLANN & visualindex \\
        \hline
        $ mAP~(\%) $ & \textbf{77.5} & 77.1  & 75.0 \\
        \Xhline{2.5\arrayrulewidth}
    \end{tabular}
    \end{scriptsize}
\end{table}

\textbf{Comparison on Image Retrieval}

Image Retrieval~\cite{JDSP10} is also a popular topic in multimedia application. It aims at retrieving the items containing the target object from a large image corpus. A typical image retrieval system is based  on the technique of bag-of-visual-words~(BOW) which mathches local features such as SIFT~\cite{Lowe:04}. And ANN search approach is heavily employed by the BOW encoding strategy. In this paper we compare our method  with the popular fast ANN approach~\cite{flann_pami_2014} for image retrieval.

We evaluate on the Oxford 5K dataset. SIFT features are extracted with gravity vector constraints~\cite{perd2009efficient} and RootSIFT~\cite{arandjelovic2012three} that  use  the square root of each component of a SIFT vector is also employed. We build the codebook of 1M visual words for BOW encoding.  
In our method, we assign 8 bits to each subspace~(k=256) and the subspace number is 8. For the fast ANN approach, we followed the setup in Fast Object Retrieval~\cite{zhong2015dsm} and visualindex~\footnote{\url{https://github.com/vedaldi/visualindex}}. In the experiment, we use mean Average Precision~(mAP) as the performance metric. 
The result is shown in Table~\ref{tb:retrieval}. We can see that our method outperforms the other two approaches.

\begin{figure*}[tbp]
    \centering
    \subfloat[]{
    \begin{minipage}{0.48\textwidth}
        \includegraphics[width=\textwidth]{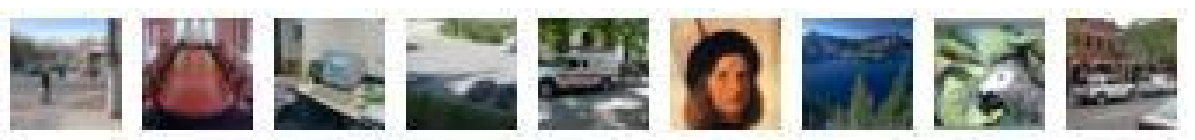}
        \includegraphics[width=\textwidth]{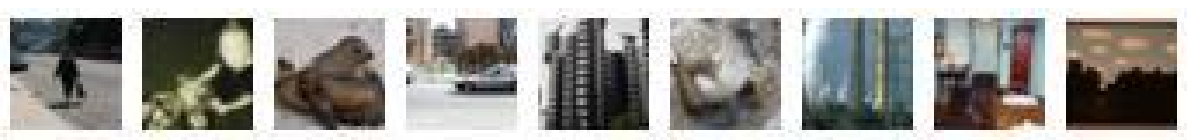}
        \includegraphics[width=\textwidth]{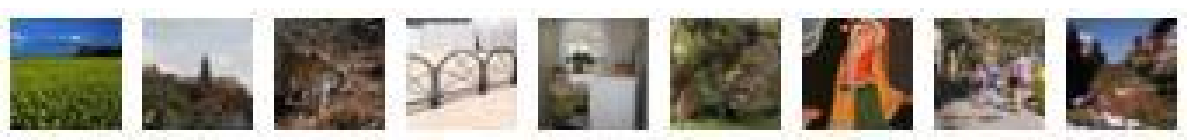}
        \includegraphics[width=\textwidth]{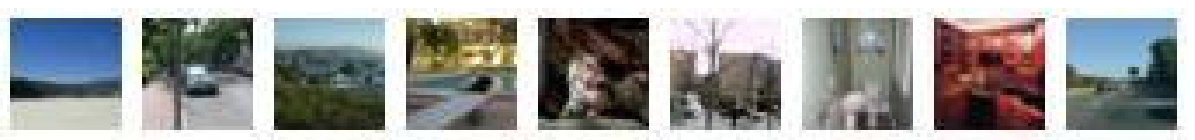}
        \includegraphics[width=\textwidth]{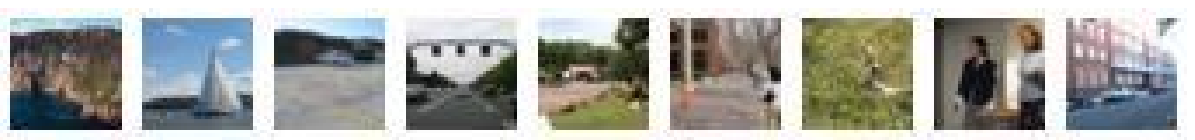}
        \includegraphics[width=\textwidth]{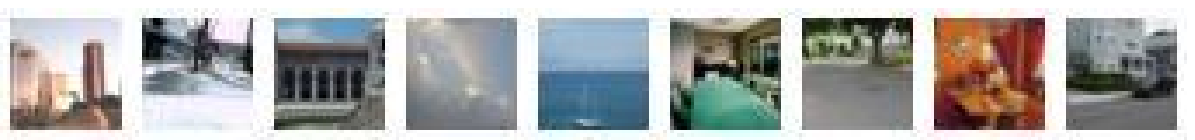}
        \includegraphics[width=\textwidth]{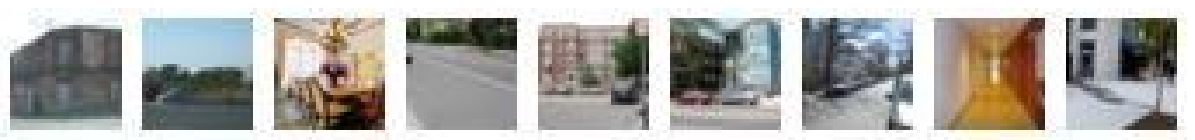}
        \includegraphics[width=\textwidth]{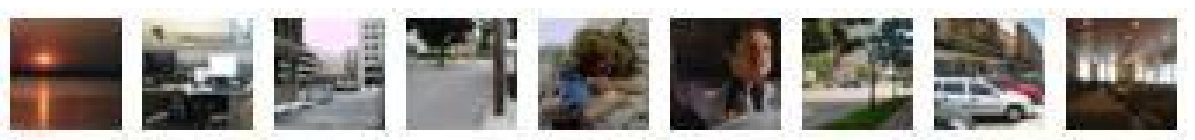}
    \end{minipage}
    \label{fig:qualitativepq}}
    \subfloat[]{
    \begin{minipage}{0.48\textwidth}
        \includegraphics[width=\textwidth]{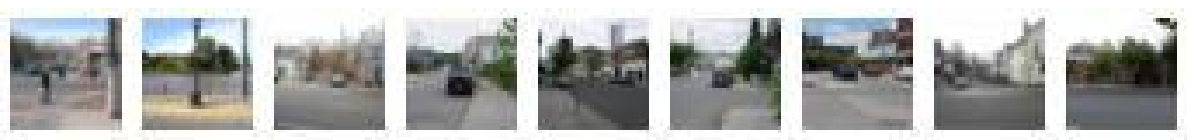}
        \includegraphics[width=\textwidth]{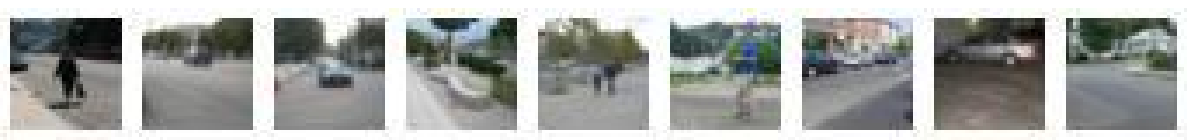}
        \includegraphics[width=\textwidth]{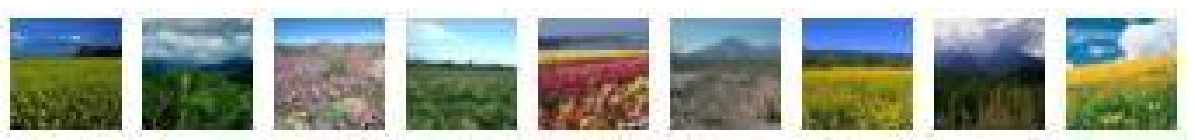}
        \includegraphics[width=\textwidth]{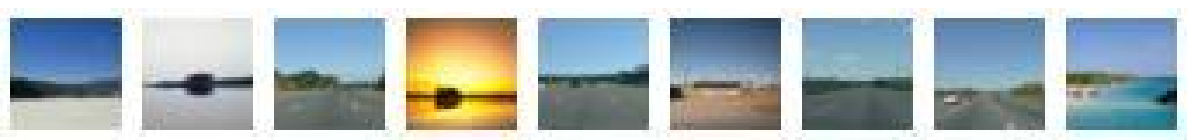}
        \includegraphics[width=\textwidth]{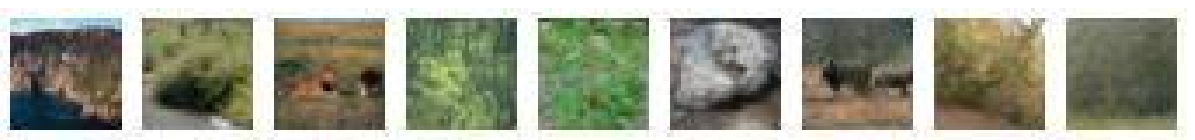}
        \includegraphics[width=\textwidth]{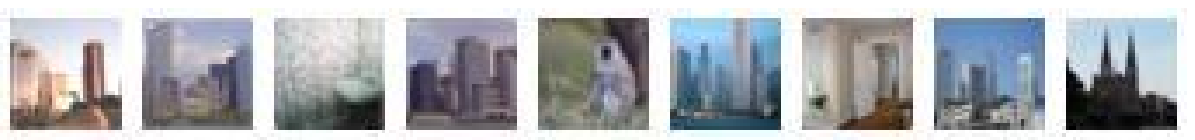}
        \includegraphics[width=\textwidth]{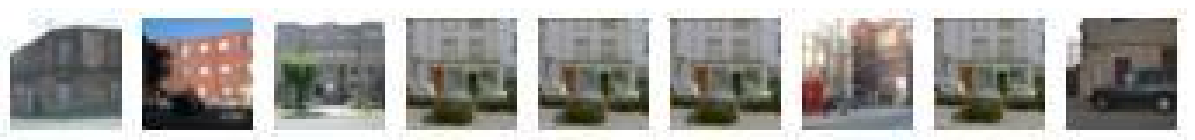}
        \includegraphics[width=\textwidth]{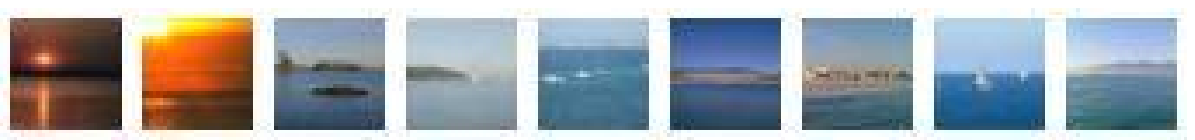}
    \end{minipage}
    \label{fig:qualitativespq}}
    \caption{Qualitative results between PQ and our proposed methods. 
        \protect \subref{fig:qualitativepq} are the results of PQ and \protect \subref{fig:qualitativespq} are the results of SPQ. For each row, the first image is a query image and the remaining are the ranking results by each method. It is obvious that our method outperforms PQ.}
    \label{fig:qualitative}%
\end{figure*}

\section{Conclusion and Future Work}
In this paper, we propose a novel Sparse Product Quantization approach to encoding high-dimensional feature vectors into sparse representation. Euclidean distance between two vectors can be efficiently estimated from their sparse product quantization using fast table lookups. We optimize the sparse representation of the data vectors by minimizing their quantization errors, making the resulting representation is essentially close to the original data in practice. We have conducted extensive experiments by evaluating the proposed Sparse Product Quantization technique for ANN search on four public image datasets, whose promising experimental results show that our method is fast and accurate, and significantly outperforms several state-of-the-art approaches with large margin.
Furthermore, the result on the image retrieval also demonstrates the efficacy of our proposed method.

Despite these promising results, some limitations and future work should be addressed. As many other soft assignment methods, the performance gain of our approach involves with extra storage requirements and computational cost inevitably. For future work, we will study how to compress the coding coefficients. Also, we will extend our technique to other tasks, such as object retrieval.

\section*{Acknowledgment}

The work was supported in part by National Natural Science Foundation of China under Grants (61103105 and 91120302).

\ifCLASSOPTIONcaptionsoff
  \newpage
\fi



\bibliographystyle{abbrv}
\bibliography{ref}  

\begin{IEEEbiography}[{\includegraphics[width=1in,height=1.25in,clip,keepaspectratio]{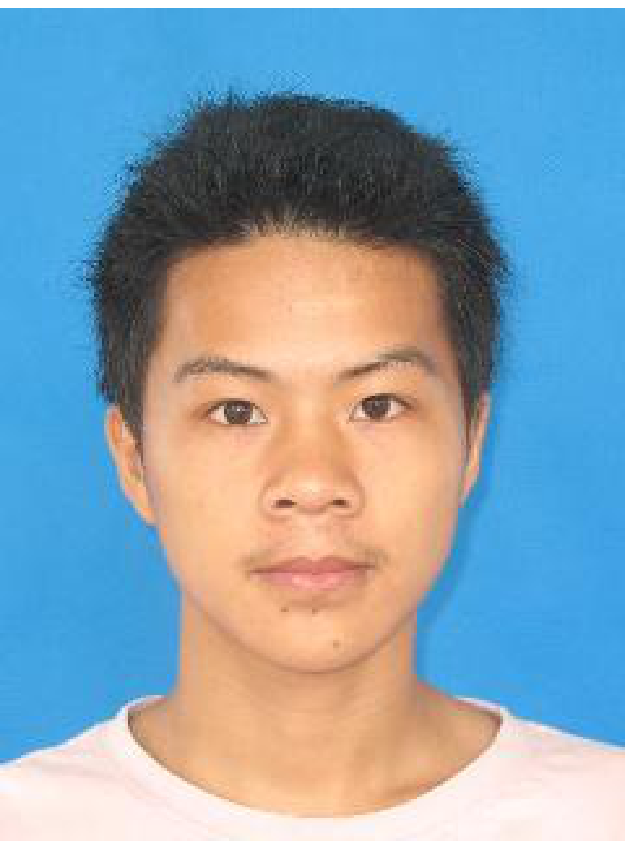}}]{Qingqun~Ning}
is currently a PhD candidate in the Department of Computer Science and Technology, Zhejiang University of China. Before that, he received the BS degree from  Northwestern Polytechnical University of China in 2011. His research interests include machine learning and computer vision, with a focus on large scale image search and object detection.
\end{IEEEbiography}
\begin{IEEEbiography}[{\includegraphics[width=1in,height=1.25in,clip,keepaspectratio]{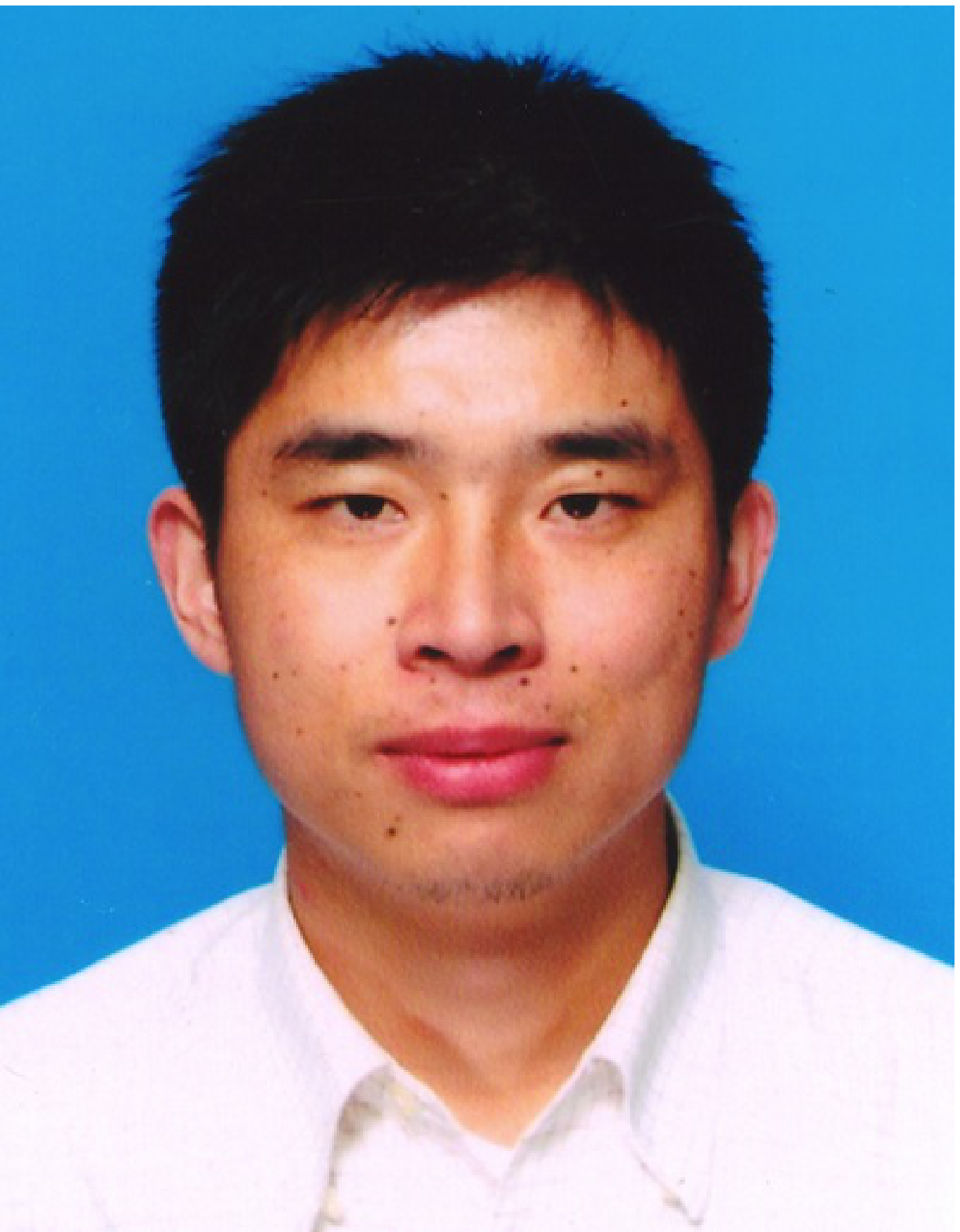}}]{Jianke Zhu}
is an Associate Professor in College of Computer Science at Zhejiang University. He received his Ph.D degree in Computer Science and Engineering from The Chinese University of Hong Kong. He was a postdoc in BIWI Computer Vision Lab at ETH Zurich. Dr. Zhu's research interests include computer vision and multimedia information retrieval. He is a member of the IEEE.
\end{IEEEbiography}

\begin{IEEEbiography}[{\includegraphics[width=1in,height=1.25in,clip,keepaspectratio]{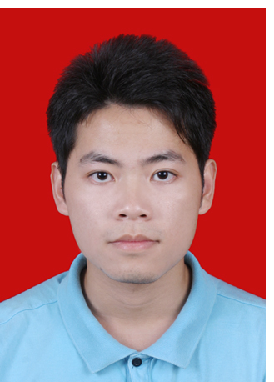}}]{Zhiyuan~Zhong}
is currently a Master student in the Department of Computer Science and Technology, Zhejiang University.
He received his Bachelor degree in Computer Science from South China University of Technology, Guangzhou, P.R. China.
His research interests include machine learning, computer vision and multimedia information retrieval.
\end{IEEEbiography}

\begin{IEEEbiography}[{\includegraphics[width=1in,height=1.25in,clip,keepaspectratio]{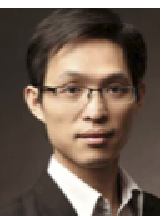}}]{Steven C.H. Hoi} 
    received the bachelor’s degree in computer science from Tsinghua University, Beijing, P.R. China, and the master’s and PhD degrees in computer science and engineering from the Chinese University of Hong Kong. He is currently an associate professor in the School of Information Systems, Singapore Management University, Singapore. His research interests include machine learning, multimedia information retrieval, web search, and data mining. He is
a member of the IEEE and ACM.
\end{IEEEbiography}

\begin{IEEEbiography}[{\includegraphics[width=1in,height=1.25in,clip,keepaspectratio]{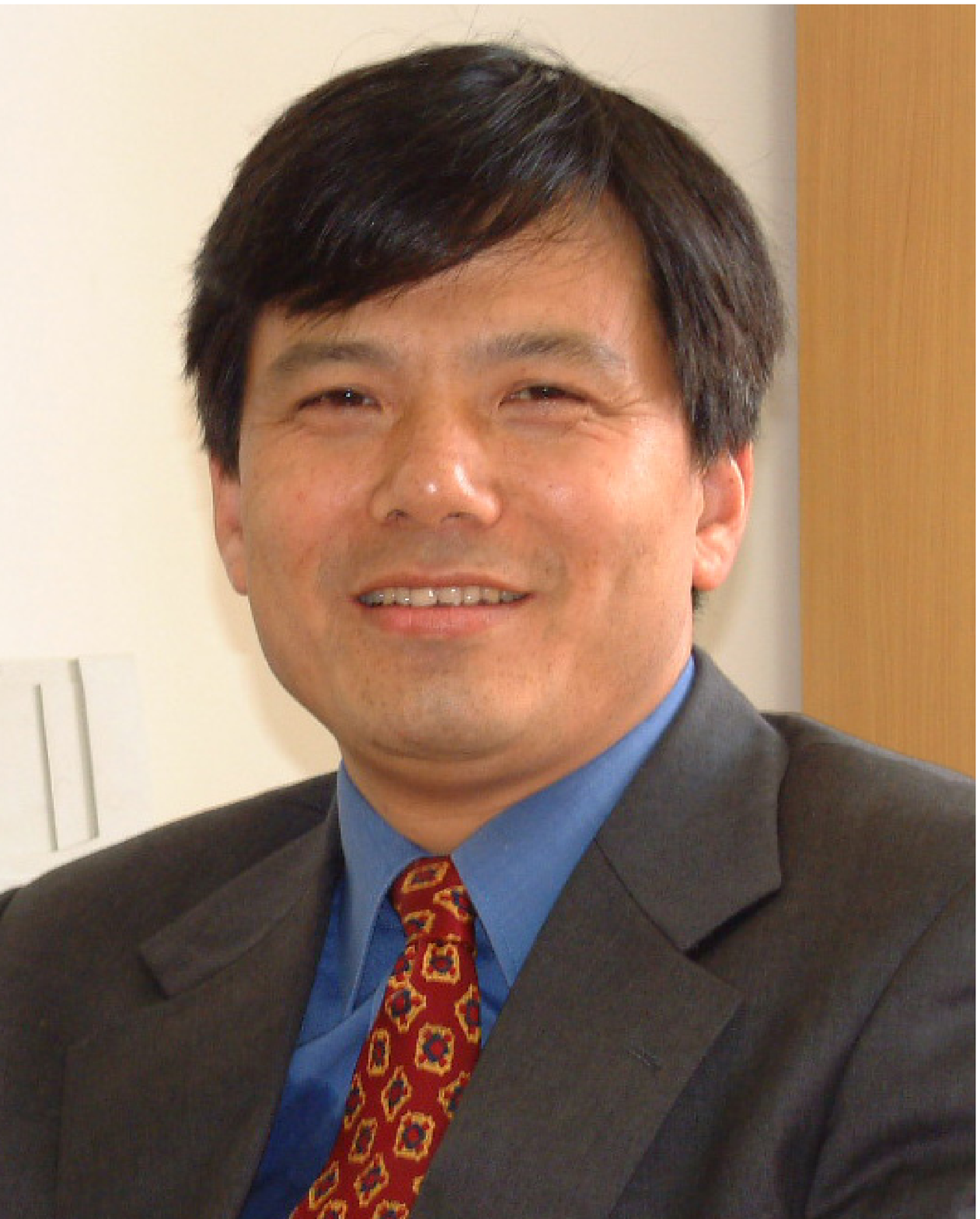}}]{Chun Chen}
 received the PhD degree from Zhejiang University in 1990. He is a professor in College of Computer Science, the Dean of College of Software, and the Director of Institute of Computer Software at Zhejiang University. His research interests include image processing, computer vision, embedded system and information retrieval. He is a member of the IEEE.
\end{IEEEbiography}



\end{document}